\RequirePackage{fix-cm}
\documentclass[twocolumn]{svjour3}
\smartqed
\usepackage{amsmath}
\usepackage{natbib}
\usepackage{bbm}
\usepackage{bm}
\usepackage{booktabs}
\usepackage{amssymb}
\usepackage{mathtools}
\usepackage{array}
\usepackage{algorithm}
\usepackage{algorithmic}
\usepackage{graphicx}
\usepackage{multirow}
\usepackage{amsfonts}
\usepackage{colortbl}
\usepackage{makecell}
\usepackage{float}
\usepackage{enumitem}
\usepackage[shortcuts]{extdash}
\usepackage{amsmath}
\usepackage{microtype}
\allowdisplaybreaks
\definecolor{darkgreen}{rgb}{0.0, 0.5, 0.0}
\definecolor{text_red}{RGB}{220,20,60}

\newcommand{\G}[1]{\textcolor{darkgreen}{$\kern 0.15em ^{#1}$}}
\newcommand{\GG}[2]{\textbf{#1}\textcolor{darkgreen}{$\kern 0.15em ^{#2}$}}
\newcommand{\GGG}[2]{\textbf{#1}\textcolor{darkgreen}{$\kern 0.15em ({#2})$}}
\newcommand{\GGGG}[2]{{#1}\textcolor{darkgreen}{$\kern 0.15em ({#2})$}}

\usepackage{pifont}

\usepackage[table,dvipsnames]{xcolor}
\usepackage{colortbl}

\newcommand{\eg}{\emph{e.g.}}

\newcommand{\etal}{\emph{et al.}}

\definecolor{remark}{rgb}{1,.5,0}
\definecolor{citecolor}{rgb}{0,0.443,0.737}
\definecolor{linkcolor}{rgb}{0.956,0.298,0.235}
\definecolor{gray}{gray}{0.5}
\definecolor{teal}{rgb}{0,0.5,0.5}
\definecolor{lightskyblue}{rgb}{0.53,0.8,0.976}

\usepackage{xspace}

\usepackage{amsmath}

\usepackage[pagebackref,breaklinks,colorlinks,bookmarks=false]{hyperref}

\colorlet{dark-blue}{blue!70!black}
\colorlet{dark-green}{green!60!black}
\colorlet{dark-red}{red!80!black}
\definecolor{mypink}{RGB}{219, 48, 122}
\hypersetup{
 colorlinks=true,
 citecolor=citecolor,
 filecolor=citecolor,
 linkcolor=linkcolor,
 urlcolor=magenta
}

\makeatletter
\renewcommand\paragraph{
  \@startsection{paragraph}
  {4}
  {\z@}
  {.5em \@plus1ex \@minus.2ex}
  {-.5em}
  {\normalfont\normalsize\bfseries}
}
\makeatother
\journalname{}
\makeatletter
\renewcommand{\makeheadbox}{}
\makeatother

\begin{document}

\title{Adaptive Causal Alignment for High-Confidence Adversarial Training}
\subtitle{}

\author{
Zhiming Luo$^{\dagger}$ \and
Kejia Zhang$^{\dagger}$ \and
Yingxin Lai \and
Junwei Wu \and
Juanjuan Weng$^*$ \and
Shaozi Li
}

\institute{
Zhiming Luo$^{\dagger}$ (Equal contribution) \at
Department of Artificial Intelligence, Xiamen University, Xiamen, China \\
\email{zhiming.luo@xmu.edu.cn}
\and
Kejia Zhang$^{\dagger}$ (Equal contribution) \at
Department of Artificial Intelligence, Xiamen University, Xiamen, China \\
\email{Kejiaz171@gmail.com}
\and
Yingxin Lai \at
Department of Artificial Intelligence, Xiamen University, Xiamen, China \\
\email{laiyingxin2@gmail.com}
\and
Junwei Wu \at Department of Computer Science, Emory University, GA, USA \\
\email{junwei.wu@emory.edu}
\and
Juanjuan Weng (Corresponding author)\at
College of Information Science and Technology, Jinan University, Guangzhou, China \\
\email{wengjj@jnu.edu.cn}
\and
Shaozi Li \at
Department of Artificial Intelligence, Xiamen University, Xiamen, China \\
\email{szlig@xmu.edu.cn}
}

\date{}

\maketitle

\begingroup
\renewcommand\thefootnote{$\dagger$}

\endgroup

\begin{abstract}
Inverse adversarial training leverages high-confidence predictions to stabilize robust learning, yet we uncover a critical paradox: high confidence often stems from overfitting to non-causal background correlations rather than intrinsic object semantics.
Our investigation reveals that visual context functions as a dual-natured signal, serving as either a necessary supportive prior or a spurious confounder.
This insight renders existing blind suppression strategies flawed, as they inevitably lead to severe Feature Loss.
To resolve this, we propose High-Confidence Causally Aligned Training (HICAT), a unified framework that establishes a Semantic Equilibrium.
Operating on a ``Measure-Debias-Align'' pipeline, HICAT integrates a Learnable Background-Bias Estimator (LBBE) to adaptively diagnose context utility.
Guided by this diagnosis, an Adaptive Debiasing mechanism performs surgical logit rectification, complemented by a geometrically grounded Foreground Logit Orthogonal Enhancement (FLOE) loss to enforce rigorous feature disentanglement.
Extensive experiments on CIFAR-10, CIFAR-100, and ImageNet-1K demonstrate that HICAT consistently improves over matched baselines across diverse architectures (CNNs and ViTs) while significantly reducing the robust generalization gap.
\end{abstract}

\section{Introduction}

Deep neural networks (DNNs) have achieved remarkable success in visual recognition~\citep{butt2025r,zhang2025tars,zhu2025obs,bai2026dice,wang2026earlytom} yet remain highly vulnerable to imperceptible adversarial perturbations~\citep{szegedy2013intriguing,fawzi2018adversarial,li2019universal}. To address these security concerns, Adversarial Training (AT) has emerged as the most effective defense paradigm~\citep{shafahi2019adversarial,kuang2024defense}. Building on this, numerous methods have been proposed to refine AT, aiming to enhance training stability, convergence, and robustness against stronger attacks~\citep{li2023wat,subramanian2024spatial,huang2023fast}.

\begin{figure*}[t]
    \centering
    \setlength{\tabcolsep}{1pt}
    \renewcommand{\arraystretch}{1}
    \begin{tabular}{@{}cc@{}}
        \includegraphics[width=0.49\linewidth]{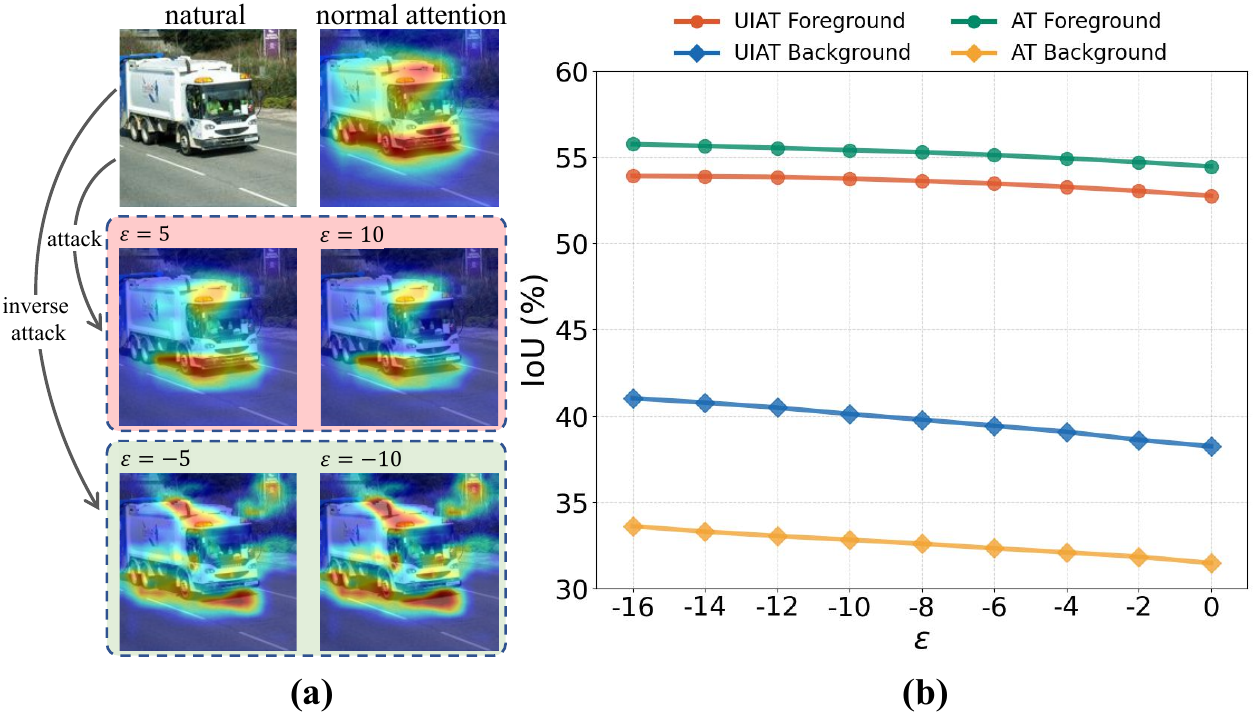} &
        \includegraphics[width=0.49\linewidth]{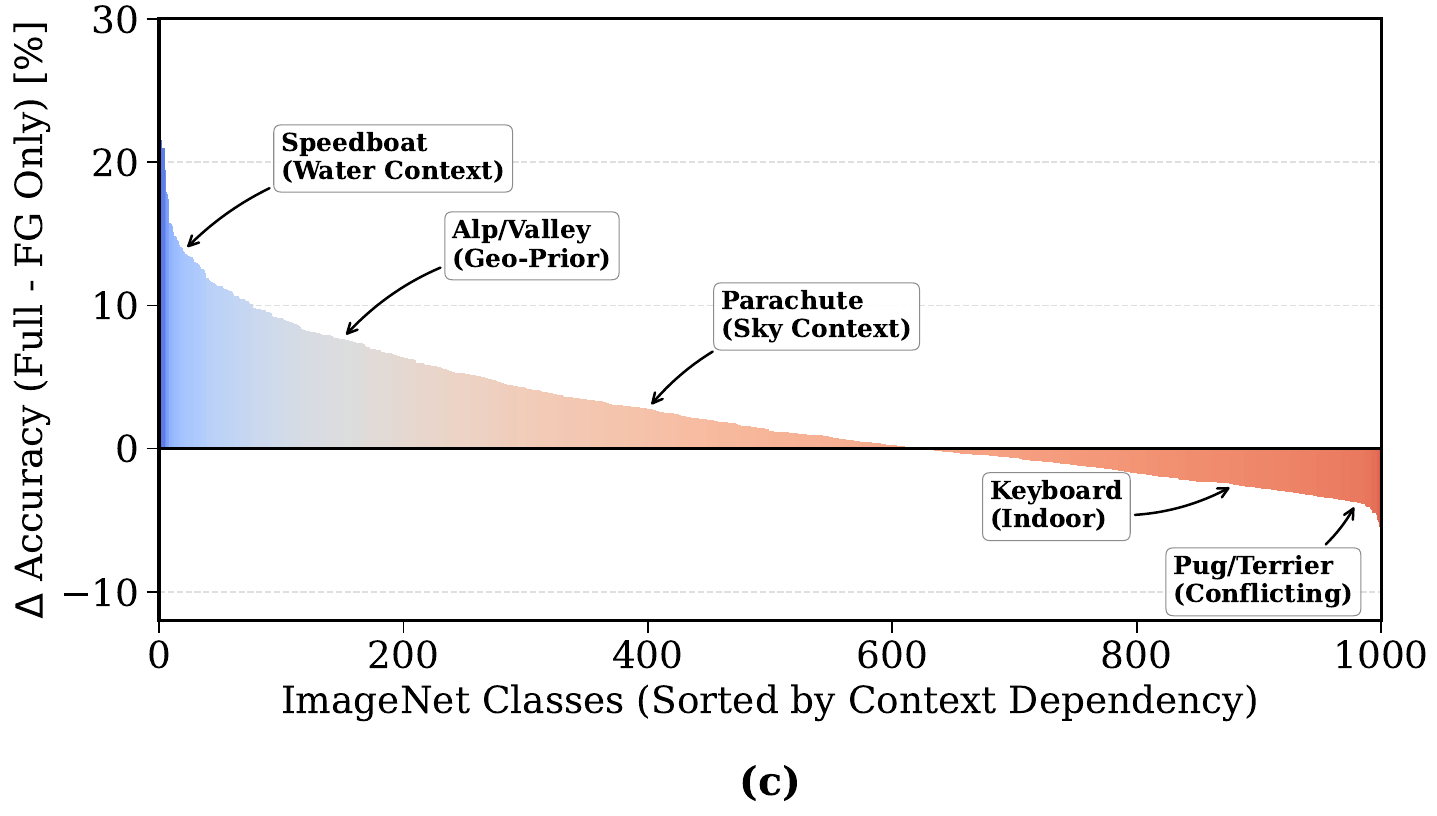}
    \end{tabular}
    \caption{
        \textbf{The Dilemma of Contextual Dependence.}
        (a) \textbf{Biased Activation:} Grad-CAM visualizations comparing feature activation under adversarial and inverse adversarial attacks. We observe that inverse attacks amplify model confidence but systematically shift attention from the foreground to the background.
        (b) \textbf{Background Alignment:} Quantitative analysis of the Intersection over Union (IoU) with background regions. The negative $\epsilon$ denotes the strength of inverse adversarial perturbations. As the inverse strength ($|\epsilon|$) increases, the UIAT model exhibits a disproportionate increase in background alignment compared to standard AT.
        (c) \textbf{Context Dependency Spectrum:} We quantify the impact of context by measuring the accuracy drop ($\Delta \text{Acc}$) on ImageNet classes after removing backgrounds using SAM masks. The results reveal a clear heterogeneity: classes on the left rely on supportive context (large drop), while those on the right are dominated by spurious context (low or negative drop).
    }
    \label{fig:motivation}
\end{figure*}

To address the vulnerabilities inherent in standard DNN models, numerous advanced AT-based techniques have been proposed~\citep{li2023wat, subramanian2024spatial, huang2023fast, li2024asam}. Prominent methods in this domain primarily focus on aligning the distributions of adversarial examples with their corresponding true classes, thereby improving classification accuracy under adversarial attacks. For instance, MART~\citep{MART} and TRADES~\citep{TRADES} aim to align the logits of adversarial examples with their corresponding natural examples to promote prediction consistency. Building upon this foundation, methods such as UIAT~\citep{UIAT} and ACR~\citep{cho2023anti} further extend the alignment strategy by generating inverse adversarial examples~\citep{salman2021unadversarial}. These inverse examples typically exhibit higher confidence than natural samples, mitigating the influence of misclassified natural data and guiding adversarial alignment toward correct, high-confidence regions.

Upon further investigation, we identify a critical limitation in inverse adversarial training: high-confidence outputs often stem from \textbf{biased feature activation}.
To rigorously validate this phenomenon, we conducted a \textbf{statistical analysis} on a subset of ImageNet~\citep{Howard_Imagenette_2019}, using the Intersection over Union (IoU) metric to measure the precise alignment between attention maps and true foreground/background regions.
As visualized in Fig.~\ref{fig:motivation}~(a) and quantified in Fig.~\ref{fig:motivation}~(b), inverse adversarial attacks consistently shift model attention away from discriminative foregrounds toward irrelevant background cues. Specifically, the Inverse Adversarial Training (UIAT) model exhibits a substantial increase in background IoU compared to standard Adversarial Training (AT)~\citep{UIAT}.
This implies that the model relies on background contextual shortcuts rather than intrinsic object features, a fundamental vulnerability we identified as \textbf{spurious correlation bias}~\citep{zhang2025towards}.

However, suppressing all background features introduces a substantial over-correction, as observed in our preliminary work DHAT~\citep{zhang2025towards}.
Background context is not merely noise; it often provides a \textbf{supportive prior} that is essential for robust reasoning~\citep{xiao2021noise}. To quantify this duality, we sample 50 images for each ImageNet class~\citep{russakovsky2015imagenet} and use SAM~\citep{kirillov2023segment} to generate foreground masks. This creates a foreground-only set for measuring the accuracy drop ($\Delta \text{Acc}$) after background removal.
The resulting \textit{Background Context Dependency Spectrum} in Fig.~\ref{fig:motivation}~(c) reveals:
\begin{itemize}
    \item \textbf{Supportive Background Context (Left Side):} A significant portion of classes experiences performance degradation when the background is removed. This indicates that even robust models rely on contextual cues, and indiscriminate background suppression in these instances leads to \textbf{Semantic Degradation}.
    \item \textbf{Spurious Background Context (Right Side):} Conversely, object-centric classes show a slight accuracy drop or even accuracy gains after background removal. This indicates that, for these classes, background information primarily manifests as \textbf{Spurious Correlation Bias}.
\end{itemize}

This empirical evidence exposes a critical dilemma: preserving all background cues can cause overfitting, while removing them can cause information loss.
Addressing this challenge requires a change from ``indiscriminate suppression'' to \textbf{Adaptive Modulation}.
We argue that genuine robustness should be founded on \textbf{Causal Alignment}, a state that balances robust foreground extraction with the preservation of supportive contextual information.

To achieve this, we extend our previous framework \textbf{DHAT}~\citep{zhang2025towards} to \textbf{High-Confidence Causally Aligned Training (HICAT)}.
HICAT incorporates a Learnable Background-Bias Estimator (LBBE) to dynamically diagnose the directional influence of contextual features, effectively distinguishing between helpful priors and harmful biases.
Informed by this diagnosis, our Adaptive Debiasing mechanism performs targeted logit recalibration.
This process suppresses spurious correlations while preserving supportive priors that align with object semantics.
Extensive experiments on CIFAR and ImageNet benchmarks demonstrate that HICAT consistently outperforms matched baselines across diverse architectures (CNNs and ViTs), effectively rectifying the semantic degradation caused by blind suppression while minimizing the robust generalization gap.

The main contributions of this work are as follows:
{
\begin{itemize}
    \item \textbf{Uncovering Contextual Duality in Adversarial Training.}
    We reveal that in adversarial training, visual context plays a dual role: serving as a necessary supportive prior for some classes while acting as spurious correlation for others. This finding challenges the premise of blind suppression strategies and necessitates an adaptive approach.

    \item \textbf{HICAT Framework.}
    We propose High-Confidence Causally Aligned Training (HICAT), which integrates a learnable estimator with adaptive logit debiasing to reduce spurious correlations while preserving semantic cues.

    \item \textbf{Strong Performance Under Matched Protocols.}
    Under matched architectures, data settings, and training schedules, HICAT delivers the strongest robustness among the compared baselines on CIFAR and ImageNet benchmarks. Crucially, it breaks the conventional clean-robust trade-off, delivering improvements in clean accuracy while enhancing adversarial robustness. Moreover, our method serves as an architecture-agnostic principle, generalizing effectively across diverse backbones (CNNs and ViTs) with a significantly reduced generalization gap.
\end{itemize}
}

\section{Related Work}
\subsection{Adversarial Attacks}

Deep neural networks (DNNs) are known to be highly sensitive to adversarial perturbations,
where a visually imperceptible noise $\delta$ added to an input $x$ can lead the model
to produce an incorrect prediction~\citep{dong2018boosting, song2024regional, guo2019simple}.
This vulnerability has been observed in a range of real-world settings such as autonomous driving,
biometric authentication, and open-world recognition~\citep{hu2023planning, yi2025d, huang2025mani, feng2024unveiling}, highlighting the importance of understanding
and defending against adversarial manipulation.
Related security concerns also appear in generated-text watermarking, where semantic-invariant attacks change the surface form while preserving meaning~\citep{ai2026pasa}.

Formally, an adversarial example $\hat{x}=x + \delta$ is generated by solving the constrained maximization:
\begin{equation}
\arg\max_{\|\delta\|_{p}\le \epsilon}
\mathcal{L}_{adv}\!\left(f_{\theta}(x+\delta),\, y\right),
\label{eq:adv_general}
\end{equation}
where $\epsilon$ bounds the perturbation under the $p$-norm and
$\mathcal{L}_{adv}$ denotes the attack objective.
This framework has motivated a variety of attack strategies differing in optimization strength and
computational cost.

\paragraph{\textbf{Gradient-based attacks.}}
The Fast Gradient Sign Method (FGSM)~\citep{FGSM} generates one-step adversaries by linearizing the loss around the input:
\begin{equation}
\hat{x}=x + \epsilon \cdot \mathrm{sign}\!\left(\nabla_{x}\mathcal{L}_{adv}(f_{\theta}(x),y)\right).
\label{eq:fgsm}
\end{equation}

To obtain stronger perturbations,
Projected Gradient Descent (PGD)~\citep{PGD_Attack} applies iterative FGSM updates followed by projection
onto the feasible set:
\begin{equation}
\hat{x}^{t+1}
=
\Pi_{\mathbb{B}(\epsilon)}
\Big[
\hat{x}^{t}
+
\alpha \cdot
\mathrm{sign}\!\left(
\nabla_{\hat{x}^{t}}
\mathcal{L}_{adv}(f_{\theta}(\hat{x}^{t}),y)
\right)
\Big],
\label{eq:pgd}
\end{equation}
where $\alpha$ is the step size and $\Pi_{\mathbb{B}(\epsilon)}$ denotes projection onto the $p$-norm ball.
The PGD is widely recognized as a universal first-order adversary and forms the foundation for most robustness evaluation setups.

\paragraph{\textbf{Optimization-based attacks.}}
The Carlini–Wagner (C\&W) attack~\citep{CW} formulates a principled optimization that jointly minimizes
perturbation magnitude and enforces misclassification through a logit-margin penalty:
\begin{equation}
\min_{\delta}\;
\|\delta\|_{2}^{2}
+
c\cdot
\max\Big(
Z(x+\delta)_{y}
-
\max_{k\neq y} Z(x+\delta)_{k}
,\;
-\kappa
\Big),
\label{eq:cw_final}
\end{equation}
where $Z(\cdot)$ denotes the neural network logits, $\kappa$ specifies the enforced misclassification margin,
and $c$ governs the strength of the penalty term.
This formulation is capable of generating highly imperceptible and targeted adversarial examples.

\paragraph{\textbf{Benchmark Attacks.}}
AutoAttack (AA) \citep{AA} has become the standard robustness evaluation protocol due to its deterministic,
parameter-free design and its use of multiple complementary attack components.
By eliminating hyperparameter tuning and ensuring repeatability, AA provides a reliable robustness estimate and
is now widely adopted as the default benchmark for assessing adversarial defenses.

\subsection{Adversarial Defense}

Adversarial defense seeks to enhance the robustness of deep neural networks against worst-case perturbations~\citep{liao2018defense, zhang2025mitigating, sriramanan2020guided}.
Among existing approaches, adversarial training is regarded as the most reliable and widely used strategy because it incorporates adversarial examples directly into the optimization process~\citep{wang2024revisiting,shrivastava2017learning,zhang2025adversarial}.

\paragraph{\textbf{Adversarial Training.}}
Classical adversarial training is commonly formulated as a min-max optimization problem~\citep{wang2021adversarial} in which the inner maximization identifies the most harmful perturbation within an $\ell_p$ ball and the outer minimization updates model parameters:
\begin{equation}
\min_{\theta}
\;
\mathbb{E}_{(x,y)\sim\mathcal{D}}
\Big[
\max_{\|\delta\|_p\le \epsilon}
\mathcal{L}_{\text{AT}}\big(f_{\theta}(x+\delta),\,y\big)
\Big].
\label{eq:at_general}
\end{equation}
Here $\mathcal{L}_{\text{AT}}$ denotes the adversarial-training objective used for the standard adversarial example, such as cross-entropy, TRADES, or MART.
A variety of extensions build upon this formulation.
TRADES~\citep{TRADES} uses a clean--adversarial consistency regularizer to balance standard accuracy and robustness. MART~\citep{MART} emphasizes misclassified examples through loss reweighting. Both methods use clean and adversarial branches, so they are useful references for computational-cost comparisons.
Adversarial Weight Perturbation improves robustness by searching for parameter perturbations that enlarge the loss landscape~\citep{AWP}.
Feature Separation and Recalibration suppresses non-robust activations in the logit space~\citep{FSR}.
Class-wise calibrated adversarial training addresses robustness imbalance across categories through class-specific weighting~\citep{CFA}.
Self-guided logit refinement stabilizes the learning trajectory by progressively refining soft label distributions~\citep{SGLR}.
Although these methods strengthen adversarial training from different perspectives, most of them implicitly assume that robustness is primarily governed by loss smoothness or feature quality, while the influence of background information on robustness is seldom addressed.

\paragraph{\textbf{Inverse Adversarial Training.}}
Another line of research investigates the use of inverse adversarial examples~\citep{UIAT}, which move the input toward regions of higher classifier confidence instead of adversarial directions.
Given an input $x$ and a confidence\-/driven objective $\mathcal{C}(\cdot)$, an inverse adversarial example $\check{x} = x-\delta$ is typically generated by solving:
\begin{equation}
\arg\min_{\|\delta\|_p\le \epsilon}
\mathcal{C}\!\left(f_{\theta}(x-\delta),\,y\right).
\label{eq:inverse_general}
\end{equation}
This formulation has been widely adopted to align adversarial features with high\-/confidence semantic targets.
Unlike prior inverse-example training that mainly uses $\check{x}$ as a high-confidence anchor~\citep{UIAT}, HICAT keeps the full inverse-example supervision signal and debiases it with a sample-wise background estimate.

However, recent studies highlight a critical vulnerability in this paradigm: high\-/confidence outputs often stem from spurious correlations rather than robust semantic features~\citep{maheronnaghsh2024robustness, varma2024ravl}.
DHAT~\citep{zhang2025towards} reveals that under inverse adversarial perturbations, the model's attention frequently shifts from causal foregrounds to irrelevant background textures~\citep{ahmad2024causal,jin2025denoising}.
To mitigate this, DHAT proposes a hard debiasing strategy comprising two components: Debiased High\-/Confidence Logit Regularization (DHLR), which subtracts background logits based on attention masks, and Foreground Logit Orthogonal Enhancement (FLOE), which enforces orthogonality between foreground and background representations.
In DHAT, the inverse-sample logit is corrected by fully subtracting the background-only component before alignment. This is equivalent to fixing the per-sample debiasing strength to one, regardless of whether the background is helpful or harmful.
While DHAT effectively suppresses spurious noise, it operates on the rigid assumption that all background context is detrimental. Consequently, its indiscriminate suppression may inadvertently discard beneficial priors~\citep{seo2022information}, leading to feature impoverishment.
HICAT replaces this fixed subtraction with an instance-adaptive gate $w(x)$ predicted by the frozen LBBE proxy. Thus, relative to DHAT, HICAT does not introduce another target-backbone pass; the extra operation is the scalar proxy gate used for sample-wise modulation.
Building on this observation, the present work introduces a learnable estimator to quantify the direction of background influence, enabling an adaptive alignment strategy that preserves helpful context while removing harmful bias.

\section{Methodology}
This section presents High\-/Confidence Causally Aligned Training (HICAT), a unified framework that strengthens model robustness by disentangling causal foreground features from spurious background correlations.
Our investigation reveals a critical paradox in existing robust training: although inverse adversarial examples effectively probe high\-/confidence decision regions, they often achieve this by over\-/exploiting non\-/causal background context rather than intrinsic object semantics.

\begin{figure}[t]
    \centering
    \includegraphics[width=\linewidth]{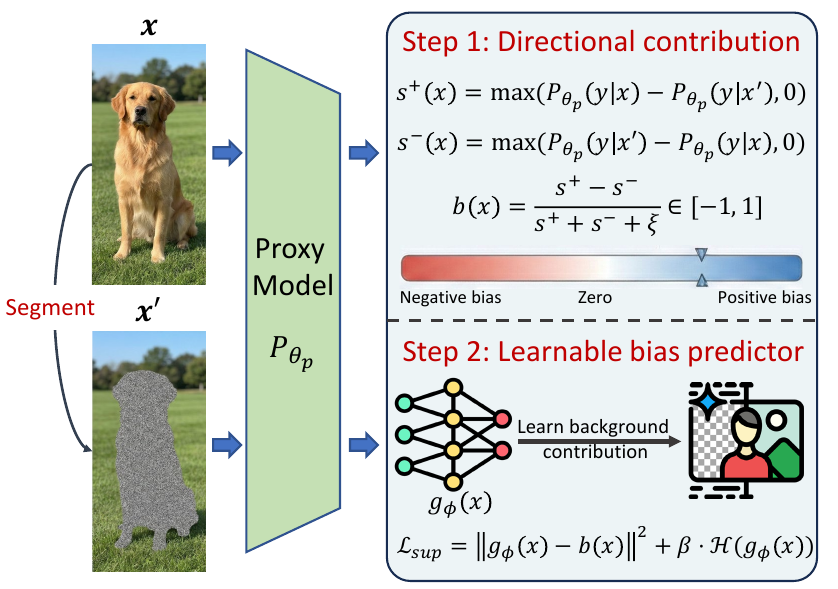}
    \caption{
    \textbf{Overview of the Learnable Background\-/Bias Estimator (LBBE).}
    The pipeline consists of three stages:
    (1) \textbf{Context Isolation}: A background\-/only variant $x'$ is generated by masking causal foreground semantics.
    (2) \textbf{Bias Quantification}: A proxy classifier computes the directional bias score $b(x)$ by comparing predictions on $x$ and $x'$, distinguishing between supportive context and spurious noise.
    (3) \textbf{Distillation}: This diagnostic signal is distilled into a lightweight predictor $g_\phi(x)$ to provide efficient guidance for the subsequent adaptive debiasing.
    }
    \label{fig:LBEE}
\end{figure}

To resolve this, HICAT operates on a \textbf{``Measure\-/Debias\-/Align''} pipeline:
\begin{itemize}
    \item \textbf{Measure (Fig.~\ref{fig:LBEE}):} We quantify the directional influence of context using a Learnable Background\-/Bias Estimator (LBBE) (Sec.~\ref{sec:lbbe}), distinguishing between helpful cues and harmful spurious biases.
    \item \textbf{Debias (Fig.~\ref{fig:Alignment}, Top Right):} Guided by this estimation, we introduce an Adaptive Debiasing mechanism (Sec.~\ref{sec:adaptive_debiasing}) that adjusts high\-/confidence logits to suppress background\-/induced activations.
    \item \textbf{Align (Fig.~\ref{fig:Alignment}, Bottom Right):} Finally, we enforce robust decision boundaries through a Debiased Logit Alignment strategy (Sec.~\ref{sec:logit_alignment}), ensuring that the model's high confidence stems from causal foreground evidence.
\end{itemize}

\subsection{Preliminaries}
\label{sec:notation}
Let $\mathcal{D}=\{(x_i, y_i)\}_{i=1}^{N}$ denote the training set, where $x_i \in \mathbb{R}^{H \times W \times C}$ represents the input and $y_i \in \{1, \dots, K\}$ is the corresponding ground\-/truth label.
We consider a deep neural network classifier $f_\theta(\cdot)$ parameterized by $\theta$. For an input $x$, it produces logits $z=f_\theta(x)\in\mathbb{R}^K$, class probabilities $p_\theta(\cdot|x)=\operatorname{softmax}(z)$, and scalar ground-truth probability $p_\theta(y|x)=[p_\theta(\cdot|x)]_y$.

To analyze and improve robustness, we consider two distinct types of perturbed examples within an $\ell_p$\-/norm constrained budget $\epsilon$:

\noindent\textbf{(1) Adversarial Examples ($\hat{x}$):}
These are standard perturbations designed to maximize the task loss $\mathcal{L}$ (\eg, cross\-/entropy), inducing model misclassification:
\begin{equation}
\begin{aligned}
\hat{x} &= x+\delta_1^\star, \\
\delta_1^\star
&= \underset{\|\delta_1\|_{\infty} \le \epsilon}{\arg\max}\,
\mathcal{L}\big(f_\theta(x+\delta_1), y\big).
\end{aligned}
\label{eq:adv_x}
\end{equation}
where $\delta_1$ represents the adversarial perturbation that pushes the input towards the decision boundary. Adversarial training aligns the model's distribution on $\hat{x}$ with the clean label $y$ to enhance robustness.

\noindent\textbf{(2) Inverse Adversarial Examples ($\check{x}$):}
Contrary to standard attacks, inverse adversarial examples are generated to \textit{minimize} the loss, effectively pushing the input further into the high\-/confidence region of the ground\-/truth class:
\begin{equation}
\begin{aligned}
\check{x} &= x-\delta_2^\star, \\
\delta_2^\star
&= \underset{\|\delta_2\|_{\infty} \le \epsilon}{\arg\min}\,
\mathcal{L}\big(f_\theta(x-\delta_2), y\big).
\end{aligned}
\label{eq:inv_x}
\end{equation}
where $\delta_2$ denotes the inverse perturbation that moves the input towards the class centroid. Since $\check{x}$ serves as a high-confidence reference for the model, we denote the logits corresponding to these inputs as $\hat{z} = f_\theta(\hat{x})$ and $\check{z} = f_\theta(\check{x})$. In our framework, $\check{z}$ serves as the primary target for debiasing and alignment.

\subsection{Learnable Background\-/Bias Estimator (LBBE)}
\label{sec:lbbe}

In visual recognition, background context plays a dual role: it acts as either a \textbf{supportive prior} that aids object identification or a \textbf{misleading distractor} that misleads the classifier under distribution shifts.
While inverse adversarial examples ($\check{x}$) effectively probe high-confidence regions, relying on them blindly risks overfitting to non-causal background patterns.
To distinguish between these two effects, we design the Learnable Background-Bias Estimator (LBBE).
As illustrated in Fig.~\ref{fig:LBEE}, this module estimates both the magnitude and the \textit{direction} of background influence via a three-stage process.

\noindent\textbf{(1) Context Isolation (Fig.~\ref{fig:LBEE}, Left).}
To evaluate the background's contribution, we first isolate contextual information by removing informative foreground semantics.
Let $M_F\in[0,1]^{H\times W}$ be a foreground probability mask derived from CAM or a segmentation model.
We generate a binary foreground indicator $\hat{M}_F$ using threshold $\tau_1$:
In our implementation, $M_F$ is generated by vanilla Grad-CAM using a RobustBench-pretrained ResNet-18 CAM source model $h_\psi$ for CIFAR-10 and ImageNet~\citep{croce2robustbench}. The CAM source model is frozen during HICAT training.
\begin{equation}
    \hat{M}_F(i,j)=\mathbb{I}\big(M_F(i,j)\ge \tau_1\big),
\end{equation}
We then construct a background\-/modified variant $x'$:
\begin{equation}
    x' = x\odot(1-\hat{M}_F) + \mathcal{N}_{\text{fg}}(\hat{M}_F),
    \label{eq:bg_only}
\end{equation}
where $\odot$ denotes element\-/wise multiplication, and $\mathcal{N}_{\text{fg}}(\cdot)$ fills the masked foreground region with Gaussian noise.
This background-only input $x'$ preserves the environmental context while removing the object, allowing us to measure the background's standalone impact on the model's decision.
This foreground-removal operation is used only for diagnosis. During adversarial training, HICAT preserves the foreground information in the inverse example and only adjusts the background-related component in Eq.~(\ref{eq:zdstar_gate}).

\noindent\textbf{(2) Bias Quantification (Fig.~\ref{fig:LBEE}, Top Right).}
In this step, we employ a trained ResNet-18 proxy classifier $p_{\theta_p}$ to compute the background's contribution.
We compare the proxy's confidence on the original input $x$ versus the background\-/modified input $x'$ for the ground\-/truth class $y$.
Here $p_{\theta_p}(\cdot|x)$ is a class-probability vector, $p_{\theta_p}(y|x)$ is its scalar entry for class $y$, and $\theta_p$ denotes the parameters of the frozen proxy classifier, distinct from the target model parameters $\theta$.
Based on this, we define the positive (helpful) and negative (harmful) contributions as:
\begin{align}
s^+(x) &= \big[p_{\theta_p}(y|x) - p_{\theta_p}(y|x')\big]_+,  \\
s^-(x) &= \big[p_{\theta_p}(y|x') - p_{\theta_p}(y|x)\big]_+.
\end{align}
where $[\cdot]_+$ is applied to one scalar confidence difference, not to a class vector.
Here, $s^+(x)$ represents the confidence gain attributed to the foreground, while $s^-(x)$ captures instances where the background alone triggers high confidence (indicating spurious dominance).
We then derive the normalized directional background bias $b(x)$:
\begin{equation}
b(x) = \frac{s^+(x) - s^-(x)}{s^+(x) + s^-(x) + \xi},
\label{eq:net_bias}
\end{equation}
where $\xi$ ensures numerical stability.

\textbf{Interpretation:} A positive score ($b(x) > 0$) implies the background is supportive or neutral. Conversely, a negative score ($b(x) < 0$) indicates that the background creates a strong spurious correlation, distracting the model from causal features.

\noindent\textbf{(3) Distillation (Fig.~\ref{fig:LBEE}, Bottom Right).}
Calculating $b(x)$ requires computationally expensive masking and proxy inference at every step, which is impractical during iterative adversarial training.
To address this, we distill this diagnostic capability into a lightweight predictor (ResNet-18) $g_\phi(\cdot)$, which is then trained to approximate the pseudo\-/label $b(x)$ on a subset of training data using the following objective:
$g_\phi: \mathbb{R}^{H\times W\times C}\rightarrow[-1,1]$ is a scalar bias regressor, not a class-logit predictor.
\begin{equation}
\mathcal{L}_{\text{sup}}
=\big\|\,g_\phi(x)-b(x)\,\big\|_2^2 \quad +\beta\cdot \mathcal{H}\big(g_\phi(x)\big),
\label{eq:lsup_new}
\end{equation}
where the first term minimizes the estimation error, and $\mathcal{H}$ denotes an entropy regularization term weighted by $\beta$ to prevent trivial solutions and ensure prediction stability.
During warm-up, $b(x)$ in Eq.~(\ref{eq:net_bias}) is computed from the proxy classifier and used as the supervision target. During adversarial training, the runtime estimate $\hat{b}(x)=g_\phi(x)$ is used for adaptive modulation.
Concretely, $g_\phi$ is trained once in a pre-adversarial warm-up stage and then frozen during adversarial training (no alternating co-optimization with $f_\theta$).
After convergence, $g_\phi(x)$ provides an efficient, sample\-/wise bias estimation. This score serves as the critical guidance for the subsequent \textbf{Adaptive Debiasing} module (Sec.~\ref{sec:adaptive_debiasing}), determining the intensity of logit recalibration.

\subsection{Adaptive Debiasing}
\label{sec:adaptive_debiasing}

\begin{figure}[t]
    \centering
    \includegraphics[width=\linewidth]{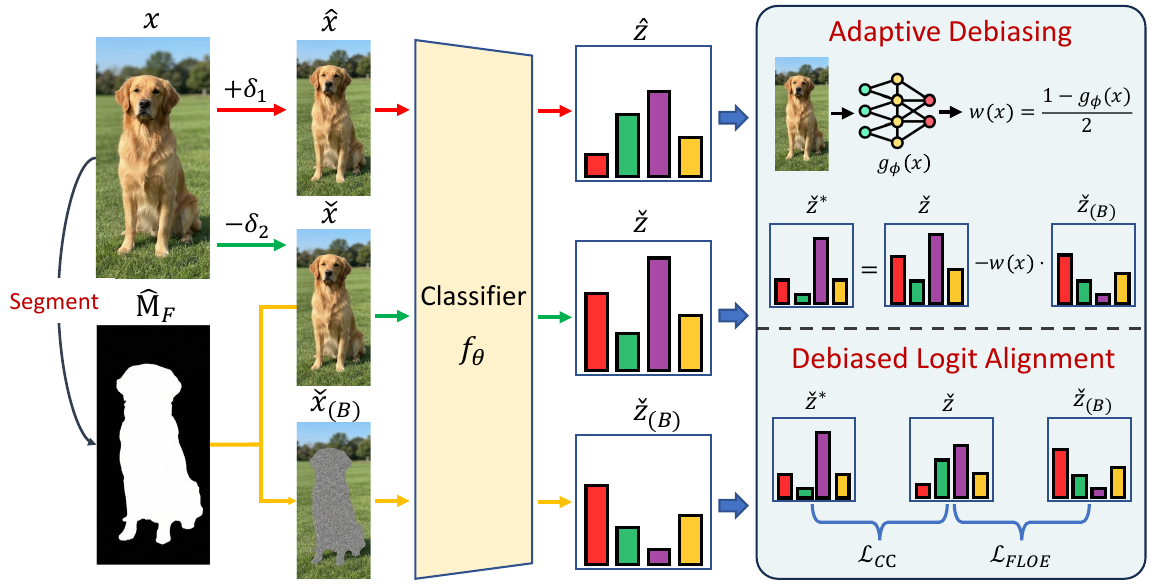}
    \caption{
    \textbf{Overview of the HICAT framework.}
    Given adversarial ($\hat{x}$) and inverse ($\check{x}$) inputs with extracted background $\check{x}_{(B)}$, the method employs two stages:
    (1) \textbf{Adaptive Debiasing}, where a bias estimator $g_\phi(x)$ computes a gating weight $w(x)$ to selectively filter spurious background activations in $\check{z}$, producing a rectified target $\check{z}^{*}$;
    (2) \textbf{Debiased Logit Alignment}, which optimizes classifier $f_\theta$ by enforcing causal consistency ($\mathcal{L}_{\text{CC}}$) and foreground-background orthogonality ($\mathcal{L}_{\text{FLOE}}$).
    }
\label{fig:Alignment}
\end{figure}

While LBBE identifies the contribution of background context, the core challenge remains: how to effectively utilize this diagnosis to rectify model predictions?
Standard adversarial training often blindly overfits to spurious background features in high-confidence regions.
To counteract this, we propose the \textbf{Adaptive Debiasing} module.
As detailed in the \textbf{Top Right panel of Fig.~\ref{fig:Alignment}}, instead of suppressing all context, this module uses the bias estimate $g_\phi(x)$ to adjust the logits and generate a debiased high-confidence representation $\check{z}^{*}$.

\noindent\textbf{(1) Background Activation Extraction (Fig.~\ref{fig:Alignment}, Input Stream).}
We utilize Inverse Adversarial Examples (IAEs), $\check{x}$, as the source of high\-/confidence supervision.
As visually depicted in the input stream of Fig.~\ref{fig:Alignment}, to ensure accurate context separation, we derive a \textbf{binary background mask} from the attention map $M$ (\eg, Grad-CAM) of the clean input $x$.
We then extract the background region $\check{x}_{(B)}$ by applying this mask to the inverse adversarial example $\check{x}$:
\begin{equation}
\begin{aligned}
[\check{x}_{(B)}]_{(i,j)}
&= \mathbb{I}\big(M_{(i,j)} < \tau_2\big)\cdot \check{x}_{(i,j)} ,
\end{aligned}
\label{eq:xb}
\end{equation}
where $\tau_2$ is a suppression threshold. The corresponding background logits are computed as $\check{z}_{(B)} = f_\theta(\check{x}_{(B)})$. These logits quantify the raw strength of contextual activations before any intervention.

\noindent\textbf{(2) Bias\-/Guided Logit Gating (Fig.~\ref{fig:Alignment}, Top Right).}
Blind subtraction of $\check{z}_{(B)}$ risks discarding helpful context. To ensure the intervention is causally grounded, we employ the distilled bias score $g_\phi(x) \in [-1, 1]$, predicted by the lightweight ResNet-18 estimator, to modulate the suppression.
We define a non\-/negative gating function $w(x)$ to transform the directional bias score into a normalized suppression weight within $[0,1]$:
\begin{equation}
\label{eq:gate_w}
w(x) = \frac{1 - g_\phi (x)}{2}.
\end{equation}
We then apply this gate to recalibrate the original high\-/confidence logits $\check{z}$:
\begin{equation}
\label{eq:zdstar_gate}
\check{z}^{*} = \check{z} - w(x)\cdot\check{z}_{(B)}.
\end{equation}
\textbf{Mechanism Analysis:} This formulation enables sample\-/adaptive debiasing:
\begin{itemize}
    \item \textit{Harmful Bias ($g_\phi \to -1$):} $w(x) \to 1$. The model identifies the background as spurious and aggressively subtracts $\check{z}_{(B)}$, purifying the logits.
    \item \textit{Helpful Context ($g_\phi \to 1$):} $w(x) \to 0$. The background is deemed supportive, so the gate closes, preserving beneficial contextual cues.
\end{itemize}
The resulting $\check{z}^{*}$ represents a rectified supervision target that retains high confidence while stripping away spurious correlations.
We first assume an approximate foreground--background separation after mask-based extraction. This condition follows prior analyses that separate foreground and background signals in object recognition~\citep{xiao2021noise}. Under this condition, foreground and background evidence are approximately separable in logit space:
\begin{equation*}
    \check{z}=z_C+z_B+\varepsilon,\quad \check{z}_{(B)}\approx z_B,
\end{equation*}
Here, $z_C$ denotes the foreground signal, and $z_B$ denotes the background signal. From Eq.~(\ref{eq:zdstar_gate}), we obtain
\begin{equation*}
    \check{z}^{*}\approx z_C+(1-w(x))z_B+\varepsilon.
\end{equation*}
Thus, the recalibrated target weakens the estimated background term while preserving the foreground term, rather than applying an unconstrained post-hoc logit shift.

\subsection{Debiased Logit Alignment}
\label{sec:logit_alignment}

With the rectified high-confidence logits $\check{z}^{*}$ obtained, the final step is to enforce these causal corrections into the model's decision boundary.
As depicted in the \textbf{Bottom Right panel of Fig.~\ref{fig:Alignment}}, we introduce a Debiased Logit Alignment strategy composed of two complementary objectives.

\noindent\textbf{(1) High-Confidence Causal Consistency ($\mathcal{L}_{\text{CC}}$).}

To transfer the ``clean" knowledge from $\check{z}^{*}$ to the robust training process, we align the standard adversarial logits $\hat{z}$ with $\check{z}^{*}$ via a KL\-/divergence constraint:
\begin{equation}
\mathcal{L}_{\text{CC}}
=\mathcal{L}_{\text{KL}}\big(\operatorname{softmax}(\check{z}^{*})||\operatorname{softmax}(\hat{z})\big),
\label{eq:lcc}
\end{equation}
Here, $\operatorname{softmax}(\cdot)$ denotes the softmax function, and $\phi$ denotes only the parameters of $g_\phi$.
Unlike standard logit pairing which might align with biased predictions, this loss anchors the adversarial prediction to the \textit{debiased} high\-/confidence target. It guides the model to rely on foreground evidence rather than distracting background.

\noindent\textbf{(2) Foreground Logit Orthogonal Enhancement ($\mathcal{L}_{\text{FLOE}}$).}
While Eq.~(\ref{eq:lcc}) aligns the distributions, residual feature entanglement may persist. To strictly enforce the separation between object and context, we propose FLOE, which imposes a geometric orthogonality constraint in the logit space:
\begin{equation}
\mathcal{L}_{\text{FLOE}}=-\left[1-|g_\phi(x)|\right]\cdot\left\|\check{z}-\operatorname{Proj}_{\check{z}_{(B)}}(\check{z})\right\|_p,
\label{eq:floe_bx}
\end{equation}
where $p$ is the norm and $\operatorname{Proj}_{\check{z}_{(B)}}(\check{z})$ denotes the projection of $\check{z}$ onto the background-logit direction.
Under the standard minimization framework, the negative sign encourages a larger orthogonal residual norm. Because this term is optimized jointly with the main adversarial loss and $\mathcal{L}_{\text{CC}}$, it acts as a regularizer that discourages background-aligned logits rather than as a standalone objective.
The weight $1 - |g_\phi(x)|$ relaxes the constraint when the background's role is clear and strengthens it when the background's role is ambiguous ($g_\phi \approx 0$).

\begin{algorithm}[t]
\caption{High-Confidence Causally Aligned Training (HICAT)}
\label{alg:hicat}
\begin{algorithmic}[1]
\REQUIRE Training set $\mathcal{D}=\{(x_i,y_i)\}$; classifier $f_\theta$;
perturbation budget $\epsilon$; background suppression threshold $\tau_2$;
hyperparameters $\lambda_1,\lambda_2$; learned bias estimator $g_\phi$.
\ENSURE Robust model parameters $\theta^*$

\FOR{each epoch}
\FOR{each mini-batch $(x,y)$}

\STATE \textbf{// 1. Generate Perturbed Examples}
\STATE $\hat{x} \leftarrow
\arg\max_{\|\delta\|_\infty \le \epsilon}\,
\mathcal{L}_{\text{AT}}(f_\theta(x+\delta_1),y)$ \COMMENT{Adversarial example}
\STATE $\check{x} \leftarrow
\arg\min_{\|\delta\|_\infty \le \epsilon}\,
\mathcal{L}_{\text{Inv}}(f_\theta(x-\delta_2),y)$ \COMMENT{Inverse adversarial example}
\STATE Compute logits: $\hat{z} \leftarrow f_\theta(\hat{x})$, $\check{z} \leftarrow f_\theta(\check{x})$.

\STATE \textbf{// 2. Measure: Quantify Background Bias (LBBE)}
\STATE Estimate directional bias score:
\STATE $\hat{b}(x) \leftarrow g_\phi(x)$ \COMMENT{Runtime bias estimate; Eq. (14) is used as warm-up supervision target}

\STATE \textbf{// 3. Debias: Adaptive Debiasing Module}
\STATE {Obtain the fixed CAM foreground heatmap $M_F$ for $x$ from the frozen CAM source model.}
\STATE Extract background activations:
\STATE {$[\check{x}_{(B)}]_{i,j} \leftarrow \mathbb{I}[(M_F)_{i,j} < \tau_2] \cdot \check{x}_{i,j}$.}
\STATE $\check{z}_{(B)} \leftarrow f_\theta(\check{x}_{(B)})$.
\STATE Compute adaptive gating weight:
\STATE $w(x) \leftarrow \frac{1 - \hat{b}(x)}{2}$.
\STATE Generate rectified high-confidence logits:
\STATE $\check{z}^{*} \leftarrow \check{z} - w(x)\cdot \check{z}_{(B)}$.

\STATE \textbf{// 4. Align: Debiased Logit Alignment}
\STATE \textbf{i. High-Confidence Causal Consistency ($\mathcal{L}_{\text{CC}}$):}
\STATE $\mathcal{L}_{\text{CC}} \leftarrow \mathcal{L}_{\mathrm{KL}}(\operatorname{softmax}(\check{z}^{*})\;\|\;\operatorname{softmax}(\hat{z}))$.
\STATE \textbf{ii. Foreground Logit Orthogonal Enhancement ($\mathcal{L}_{\text{FLOE}}$):}
\STATE {$\mathcal{L}_{\text{FLOE}} \leftarrow -(1-|\hat{b}(x)|)\cdot
\left\|
\check{z}-\frac{\langle \check{z},\check{z}_{(B)}\rangle}
{\|\check{z}_{(B)}\|_2^2}\check{z}_{(B)}
\right\|_p$.}

\STATE \textbf{// 5. Optimization}
\STATE $\mathcal{L}_{\text{HICAT}} \leftarrow \mathcal{L}_{\text{AT}}(\hat{z},y)
+ \lambda_1\cdot\mathcal{L}_{\text{CC}}
+ \lambda_2\cdot\mathcal{L}_{\text{FLOE}}$.
\STATE $\theta \leftarrow \theta - \eta\cdot\nabla_\theta \mathcal{L}_{\text{HICAT}}$.

\ENDFOR
\ENDFOR

\RETURN $\theta^*$
\end{algorithmic}
\end{algorithm}

\subsection{Training loss function}
\label{sec:opt}

The proposed framework, HICAT, integrates these components into a unified adversarial training objective:
\begin{equation}
\label{eq:hicat}
\mathcal{L}_{\text{Total}}
= \mathcal{L}_{\text{AT}}(\hat{z},y)
+ \lambda_1 \cdot \mathcal{L}_{\text{CC}} + \lambda_2 \cdot \mathcal{L}_{\text{FLOE}} .
\end{equation}
Here $\mathcal{L}_{\text{AT}}$ is the method-specific adversarial-training loss applied to the standard adversarial example $\hat{x}$, e.g., TRADES or MART.
The weights $\lambda_{1}$ and $\lambda_{2}$ balance the two added terms.
This formulation combines adaptive debiasing with adversarial training, so robustness is learned from causal feature alignment rather than spurious correlation fitting. Algorithm~\ref{alg:hicat} summarizes the training procedure.

\subsection{Theoretical Justification: Geometric Interpretation and Adaptive Regulation}
\label{sec:floe_theory}

\begin{figure}[t]
    \centering
    \includegraphics[width=0.75\linewidth]{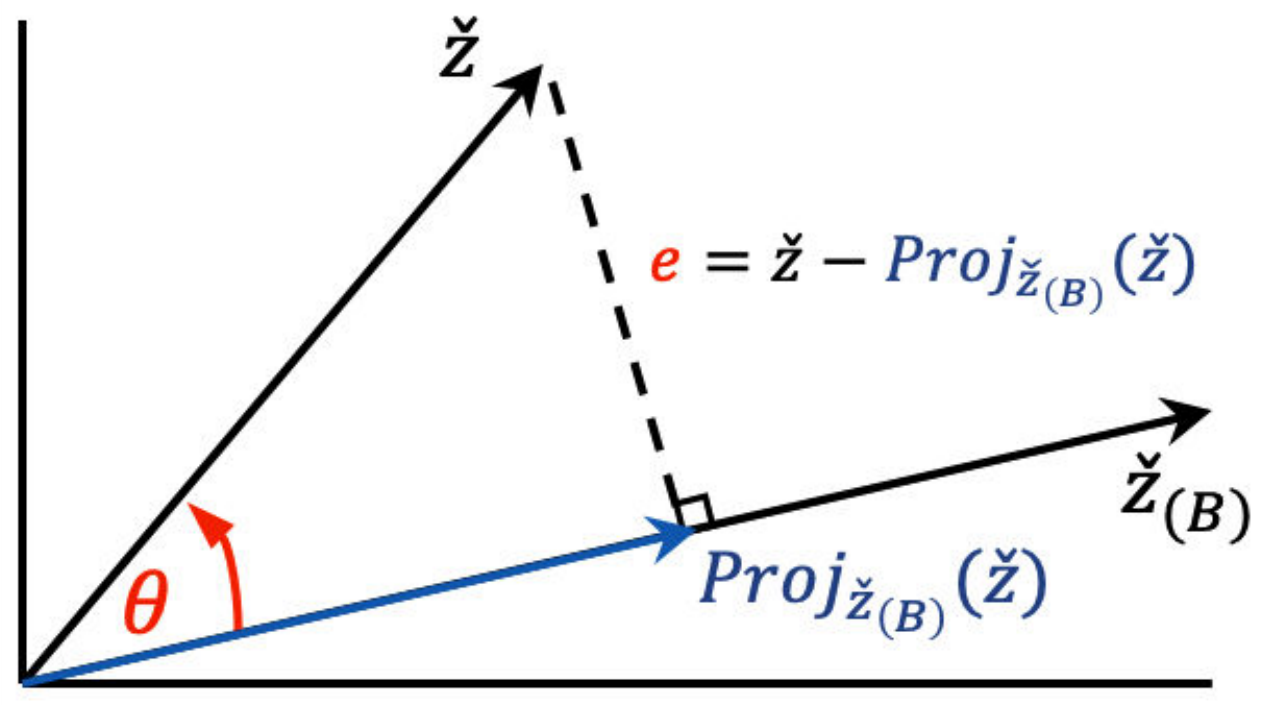}
    \caption{\textbf{Geometric Interpretation.} FLOE decomposes logits $\check{z}$ into a background projection and an orthogonal residual $e$. Encouraging a larger $\|e\|_p$ under the full objective drives the representation away from spurious bias, preserving semantics.}
    \label{fig:projection}
\end{figure}

In this section, we provide a geometric analysis of FLOE to demonstrate how it integrates feature disentanglement with uncertainty-aware modulation (Fig.~\ref{fig:projection}).

\noindent \textbf{Geometric Decomposition.}
To quantify semantic purity, we decompose the high-confidence logits $\check{z} \in \mathbb{R}^{C}$ into a background projection and an \textit{orthogonal residual} $e$ relative to background logits $\check{z}_{(B)}$, where $e$ denotes the residual after removing the background projection and $\operatorname{Proj}_{u}(v)=\langle v,u\rangle u/\|u\|_2^2$.
Under controlled logit energy, this decomposition shows that increasing the orthogonal residual reduces the relative contribution of the background projection, moving the representation away from the background-aligned direction and helping separate causal semantics from spurious correlations.
Equivalently, when logit energy is controlled,
\begin{equation}
\|\check{z}\|_2^2=\|\operatorname{Proj}_{\check{z}_{(B)}}(\check{z})\|_2^2+\|e\|_2^2,
\end{equation}
so increasing $\|e\|_2$ reduces the relative contribution of the background projection.

\noindent \textbf{Adaptive Uncertainty Regulation.}
Since strict orthogonality may discard helpful priors, HICAT uses a weight $\omega(x) = 1 - |g_\phi(x)|$ to control the FLOE strength. The final FLOE term is:
\begin{equation}
\mathcal{L}_{\mathrm{FLOE}}=-\underbrace{\left[1-|g_\phi(x)|\right]}_{\text{Uncertainty Regulator}}\cdot\underbrace{\|e\|_p}_{\text{Orthogonal Residual Norm}}.
\label{eq:floe_final}
\end{equation}
The weight $\omega(x)$ adjusts the constraint for each sample:
\begin{itemize}
    \item \textbf{Definitive Context ($|g_\phi| \approx 1$):} $\omega(x) \to 0$. This relaxes the constraint, allowing the model to integrate helpful priors ($g_\phi \to 1$) or rely on the explicit subtraction for known biases.
    \item \textbf{Ambiguous Context ($g_\phi \approx 0$):} $\omega(x) \to 1$. When the background's role is uncertain, the objective enforces maximum separation as a safety mechanism to prevent overfitting to ambiguous noise.
\end{itemize}

\begin{figure}[t]
    \centering
    \includegraphics[width=\linewidth]{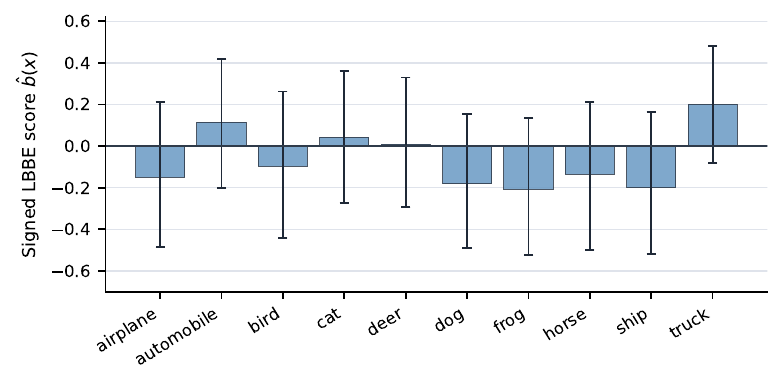}
    \caption{CIFAR-10 class-level mean background-bias scores for sample-wise and class-wise background modulation, with error bars showing within-class differences.}
    \label{fig:classwise_lbbe_diagnostic}
\end{figure}

\noindent \textbf{Sample-wise vs. Class-wise Adjustment.}
Fig.~\ref{fig:classwise_lbbe_diagnostic} reports the class-level average background-bias score and within-class differences on CIFAR-10. The figure shows that samples from the same class can have different background-bias scores. Therefore, a single class-wise weight cannot accurately adjust all samples in that class. In the class-wise variant, we compute $\bar{b}_y$ by averaging LBBE scores over all training samples in class $y$, then replace the sample-wise $\hat{b}(x)$ with $\bar{b}_y$ in CC and FLOE. Thus CC uses $w_y=(1-\bar{b}_y)/2$, and FLOE uses weight $1-|\bar{b}_y|$. Table~\ref{tb:classwise_lbbe_mean} further shows that class-wise weighting improves over fixed CC/FLOE weights (AA: 53.73\% vs.\ 52.99\%) but remains below sample-wise HICAT (54.56\%), with a larger robust gap (3.22\% vs.\ 2.86\%).

\section{Experiments}

\subsection{Experimental Setups}

\subsubsection{Architectures}
To comprehensively evaluate the universality of \textsc{HICAT}, we conduct experiments across a broad range of backbones with different inductive biases.
Our experiments include widely used convolutional networks, including ResNet\-/18, ResNet\-/50~\citep{ResNet}, VGG\-/16~\citep{VGG}, WRN28\-/10~\citep{WRN_NET}, and Inception\-/V3~\citep{Inception_NET}.
In addition, we extend our evaluation to the transformer\-/based ViT\-/B/16~\citep{dosovitskiy2020image} to verify that our method maintains robust performance across different architectural paradigms.

\subsubsection{Datasets}

\paragraph{\textbf{Datasets.}}
We evaluate our method on three widely used image classification benchmarks:
CIFAR-10 and CIFAR-100~\citep{CIFAR}, which contain 50,000 training and 10,000 test images at 32×32 resolution from 10 and 100 object categories, respectively, and ImageNet-1K~\citep{russakovsky2015imagenet}, which consists of 1.28M training and 50K validation images at high resolution (typically 224×224) across 1,000 classes.
Together, these datasets span both low-resolution and large-scale visual domains, enabling a comprehensive evaluation of robustness and generalization.

All CIFAR-10 and CIFAR-100 experiments are conducted on single NVIDIA RTX 4090 GPU per model, while ImageNet-1K experiments use two NVIDIA H100 GPUs due to the higher computational and memory requirements. This setup is sufficient to reproduce the full training schedule and ensures stable adversarial optimization across all settings.

\subsubsection{Evaluation Metrics}

\paragraph{\textbf{Adversarial Robustness.}}
We evaluate model robustness using {Robust Accuracy}, defined as the percentage of test samples that correctly classified under adversarial attacks.
The evaluations are conducted under $\ell_\infty$ constraints using PGD~\citep{PGD_Attack}, C\&W~\citep{CW}, and AutoAttack (AA)~\citep{AA}.
Notably, AA integrates an ensemble of strong attacks (APGD-DLR, APGD-CE, FAB~\citep{FAB}, and Square Attack~\citep{Square}) to provide a reliable estimate of  worst-case robustness.

\paragraph{\textbf{Robust Generalization.}}
We further measure the {robust generalization gap}, defined as the difference between robust accuracy on the training and test sets.
For a given attack $\mathcal{A}$, the gap is computed as:
\begin{equation}
\text{Gap}_{\mathcal{A}}
=
\text{RobAcc}_{\text{train}}^{\mathcal{A}}
-
\text{RobAcc}_{\text{test}}^{\mathcal{A}}.
\end{equation}
A smaller gap indicates better robust generalization and reduced robust overfitting during adversarial training.
Unless otherwise stated, we report the robust gap using PGD-10, which serves as a standard and computationally stable choice for evaluating robustness consistency across training and test data.

\subsubsection{Implementation Details}
\paragraph{\textbf{CIFAR-10 and CIFAR-100.}}
To ensure fair comparison across all methods, we adopt a unified training configuration on CIFAR-10 and CIFAR-100~\citep{CIFAR}.
All models are optimized using SGD with a momentum of 0.9 and a weight decay of $5\times10^{-4}$.
The initial learning rate is set to 0.1 and decayed by a factor of 0.1 at the 80th and 90th epochs over a total of 100 training epochs.

During adversarial training, we generate perturbations using a 10-step $\ell_\infty$-PGD attack with a budget of $\epsilon=8/255$ and a step size of $\alpha=2/255$.
For producing inverse adversarial examples required by our method, we follow the same attack configuration as the UIAT baseline~\citep{UIAT} to ensure fairness.
For Vision Transformers (ViTs), we align our setup with the training strategy proposed by Mo \etal~\citep{mo2022adversarial} to ensure training stability and convergence.

\paragraph{\textbf{ImageNet-1K.}}
\noindent
For ImageNet-1K, we adopt the training strategy~\citep{singh2023revisiting} tailored for large-scale adversarial robustness. Specifically, models are trained for 50 epochs using adversarial perturbations generated via 2-step PGD. This configuration is selected to balance computational efficiency with training stability.

\subsubsection{Hyperparameter Settings}
We follow the general configuration protocol established in DHAT~\citep{zhang2025towards}, while introducing subtle changes required by our framework. Unless otherwise stated, we set $\lambda_1=\lambda_2=1.0$ and $\tau_1=\tau_2=0.5$ across all datasets. The thresholds define the foreground mask for LBBE and the background mask for adaptive debiasing, respectively.

For foreground probability estimation, we use vanilla Grad-CAM with RobustBench-pretrained ResNet-18 CAM source models for CIFAR-10 and ImageNet~\citep{croce2robustbench}. The CAM source model is frozen during HICAT training. The heatmap is thresholded by $\tau_1$ for the foreground mask used in LBBE and by $\tau_2$ for the background mask used in adaptive debiasing.
The LBBE proxy is trained once \emph{before} adversarial training, as summarized in Table~\ref{tb:lbbe_warmup_rtx4090}. During adversarial training, both the CAM source model and $g_\phi$ are frozen; HICAT only uses forward inference of $\hat{b}(x)=g_\phi(x)$. No alternating or joint optimization with $f_\theta$ is performed.
We use a unified hardware setup for each comparison table and report the corresponding device in the caption/text.

\begin{table*}[t]
        \begin{center}
        \caption{
        Adversarial robustness (\%) and robust generalization gap (\%) across CIFAR-10, CIFAR-100, and ImageNet-1K.
        All methods are trained under a unified and fully reproducible training protocol to ensure fairness.
        The \textbf{bold} numbers indicate the best performance.
        }
        \resizebox{\textwidth}{!}
        {
            \begin{tabular}{l*{6}{>{\centering\arraybackslash}p{1.8cm}}*{1}{>{\centering\arraybackslash}p{2.1cm}}}
            \toprule
            \textbf{Net: WRN28-10} & \multicolumn{6}{c}{Attack~($\epsilon=4/255$)} & \multirow{2}{*}{Robust Gap$\downarrow$} \\
            \cmidrule{2-7}
            \textbf{ImageNet-1K}
             & Clean$\uparrow$
             & PGD-10$\uparrow$
             & PGD-20$\uparrow$
             & PGD-50$\uparrow$
             & C\&W$\uparrow$
             & AA$\uparrow$
             & \\
            \midrule
            MART~\citep{MART}   &62.31&43.56&42.89&42.54&40.62&38.94&13.76\\
            AWP~\citep{AWP}     &64.25&45.13&44.67&44.21&41.96&40.02&12.82\\
            FSR~\citep{FSR}     &64.70&44.31&43.82&43.45&41.23&39.30&14.17\\
            CFA~\citep{CFA}     &65.38&44.87&44.32&43.96&41.58&39.65&15.49\\
            UIAT~\citep{UIAT}   &62.64&45.29&44.85&44.57&42.13&40.18&14.68\\
            SGLR~\citep{SGLR}   &63.46&44.52&44.05&43.78&41.36&39.47&19.72\\
            DHAT~\citep{zhang2025towards}        &65.90&46.83&46.42&46.11&43.25&41.70&9.53\\
            \textbf{HICAT~(ours)}   &\textbf{67.12}&\textbf{47.91}&\textbf{47.52}&\textbf{47.21}&\textbf{44.60}&\textbf{42.85}&\textbf{8.72}\\
            \midrule
            \textbf{Net: WRN28-10} & \multicolumn{6}{c}{Attack~($\epsilon=8/255$)} &\multicolumn{1}{c}{\multirow{2}{*}{Robust Gap$\downarrow$}}\\
            \cmidrule{2-7}
            \textbf{CIFAR-10}&Clean$\uparrow$& PGD-10$\uparrow$&PGD-20$\uparrow$&PGD-50$\uparrow$&C\&W$\uparrow$&AA$\uparrow$&\\
            \midrule
            MART~\citep{MART}     &82.99&56.25&55.48&55.45&52.26&50.67&9.52\\
            AWP~\citep{AWP}       &82.67&57.80&57.21&57.07&54.82&51.90&6.90\\
            FSR~\citep{FSR}       &82.92&56.69&55.94&55.51&53.93&51.74&7.42\\
            CFA~\citep{CFA}       &84.43&57.87&56.90&56.64&54.60&51.85&10.36\\
            UIAT~\citep{UIAT}     &82.94&58.66&58.12&58.05&54.11&52.17&7.92\\
            SGLR~\citep{SGLR}     &85.76&57.53&56.91&56.66&54.28&52.07&9.38\\
            DHAT~\citep{zhang2025towards}          &83.95&60.49&59.95&59.87&55.27&53.10&3.51\\
            \textbf{HICAT~(ours)} &\textbf{85.92}&\textbf{61.87}&\textbf{61.42}&\textbf{61.11}&\textbf{56.59}&\textbf{54.56}&\textbf{2.86}\\
            \midrule
            \textbf{Net: WRN28-10} & \multicolumn{6}{c}{Attack~($\epsilon=8/255$)} &\multicolumn{1}{c}{\multirow{2}{*}{Robust Gap$\downarrow$}}\\
            \cmidrule{2-7}
            \textbf{CIFAR-100}&Clean$\uparrow$&PGD-10$\uparrow$&PGD-20$\uparrow$&PGD-50$\uparrow$&C\&W$\uparrow$&AA$\uparrow$&\\
            \midrule
            MART~\citep{MART}   &54.69&32.06&31.90&31.88&28.77&27.25&9.96\\
            AWP~\citep{AWP}     &57.94&34.01&33.75&33.72&30.74&28.90&7.87\\
            FSR~\citep{FSR}     &57.48&32.93&32.30&32.26&29.16&27.04&7.84\\
            CFA~\citep{CFA}     &60.92&33.10&32.56&32.41&30.49&28.04&10.47\\
            UIAT~\citep{UIAT}   &57.65&34.27&33.91&33.85&30.97&29.03&11.70\\
            SGLR~\citep{SGLR}   &61.02&33.43&32.98&32.82&30.72&28.50&15.67\\
            DHAT~\citep{zhang2025towards}                &59.14&35.82&35.33&35.02&31.72&30.17&4.24\\
            \textbf{HICAT~(ours)} &\textbf{62.38}&\textbf{37.29}&\textbf{36.82}&\textbf{36.71}&\textbf{33.15}&\textbf{31.24}&\textbf{3.12}\\
            \bottomrule[1.5pt]
            \end{tabular}
        }
        \label{st:SOTA_Comparison}
        \end{center}
\end{table*}

\subsection{Main Results on Benchmark Datasets}
\label{sec:main_results}

We report the performance of HICAT on CIFAR-10, CIFAR-100, and ImageNet-1K. Table~\ref{st:SOTA_Comparison} summarizes the results in terms of clean accuracy, robustness, and generalization gaps. Under matched architectures, training schedules, and data protocols, HICAT consistently outperforms the compared baselines, demonstrating the effectiveness and universality of our causal alignment strategy.

\begin{itemize}
    \item \textbf{Strong Robustness Gains Under Matched Protocols.}
    HICAT achieves the strongest defense among the compared methods across all benchmarks, improving AutoAttack (AA) to 54.56\% on CIFAR-10 (+1.46\% over DHAT) and 42.85\% on ImageNet-1K (+1.15\% over DHAT).
    This improvement stems from adaptive debiasing, which removes spurious correlations while preserving useful context. HICAT grounds high confidence in object semantics and remains effective from CIFAR-scale images to ImageNet.

    \item \textbf{Mitigating the Clean-Robust Trade-off via Contextual Diagnosis.}
    Notably, HICAT enhances robustness without sacrificing clean accuracy, achieving 85.92\% clean accuracy on CIFAR-10 and 67.12\% on ImageNet (+1.22\% over DHAT).
    This advantage arises from the LBBE module, which diagnoses the utility of contextual cues. By preserving supportive priors rather than indiscriminately discarding them, HICAT avoids feature impoverishment, which is particularly vital for resolving semantic ambiguity in natural images.

    \item \textbf{Superior Generalization via Orthogonal Disentanglement.}
    HICAT demonstrates the smallest robust generalization gap, reducing it from 3.51\% (DHAT) to 2.86\% on CIFAR-10 and from 9.53\% (DHAT) to 8.72\% on ImageNet.
    This reduction confirms that FLOE mitigates robust overfitting. By enforcing foreground-background orthogonality, HICAT reduces memorization of training-specific background noise and improves test robustness.
\end{itemize}

\begin{table*}[t]
    \begin{center}
    \caption{
    Comparison of robustness~(\%) and robust generalization gap~(\%) across CNN architectures on CIFAR-10 under $\ell_\infty$ perturbations ($\epsilon=8/255$).
    The \textbf{bold} numbers indicate the best performance.
    }
    \label{st:different_net}
    \resizebox{\textwidth}{!}
    {
        \begin{tabular}{lccccccccc}
        \toprule[1.5pt]
        \textbf{CIFAR-10} & \multicolumn{3}{c}{ResNet-50} & \multicolumn{3}{c}{VGG-16} & \multicolumn{3}{c}{Inception-V3}\\
        \cmidrule(lr){2-4} \cmidrule(lr){5-7} \cmidrule(lr){8-10}
        & Clean$\uparrow$ & AA$\uparrow$ & Robust Gap$\downarrow$
        & Clean$\uparrow$ & AA$\uparrow$ & Robust Gap$\downarrow$
        & Clean$\uparrow$ & AA$\uparrow$ & Robust Gap$\downarrow$\\
        \midrule
        MART~\citep{MART}     &75.05&49.05&4.57&66.72&44.27&3.60&76.05&49.59&5.91\\
        AWP~\citep{AWP}       &75.59&50.80&3.91&67.14&45.97&3.47&76.74&50.46&4.07\\
        FSR~\citep{FSR}       &78.38&50.59&3.97&69.59&44.90&3.30&77.28&50.02&4.73\\
        CFA~\citep{CFA}       &77.88&50.64&4.27&70.45&45.53&4.66&76.70&49.97&4.46\\
        UIAT~\citep{UIAT}     &78.42&51.00&2.99&68.46&45.27&3.57&75.19&51.23&3.70\\
        SGLR~\citep{SGLR}     &80.39&50.55&5.21&70.90&45.04&3.94&80.62&50.73&5.00\\
        DHAT~\citep{zhang2025towards} &79.94&51.92&1.85&70.01&46.40&1.34&77.85&51.81&1.92\\
        \textbf{HICAT (ours)} &\textbf{81.62}&\textbf{53.24}&\textbf{1.72}&\textbf{71.18}&\textbf{47.03}&\textbf{1.21}&\textbf{78.93}&\textbf{53.10}&\textbf{1.64}\\
        \bottomrule[1.5pt]
        \end{tabular}
    }
    \end{center}
\end{table*}

\begin{table}[t]
    \centering
    \caption{Comparison on ImageNet with the ViT-B/16 backbone under AutoAttack ($\epsilon=8/255$).}
    \label{tb:imagenet_vit_b16}
    \resizebox{\linewidth}{!}{
    \begin{tabular}{lccc}
        \toprule
        \textbf{Method} & \textbf{Clean (\%)} & \textbf{AA (\%)} & \textbf{Robust Gap (\%)} \\
        \midrule
        MART~\citep{MART} & 64.28 & 44.71 & 16.70 \\
        AWP~\citep{AWP} & 67.51 & 46.23 & 14.31 \\
        FSR~\citep{FSR} & 67.22 & 45.33 & 12.40 \\
        CFA~\citep{CFA} & 66.90 & 44.64 & 15.16 \\
        UIAT~\citep{UIAT} & 66.87 & 46.38 & 17.56 \\
        SGLR~\citep{SGLR} & 67.14 & 47.30 & 15.07 \\
        DHAT~\citep{zhang2025towards} & 68.59 & 48.91 & 11.82 \\
        \textbf{HICAT (ours)} & \textbf{69.77} & \textbf{50.24} & \textbf{9.79} \\
        \bottomrule
    \end{tabular}}
\end{table}

\begin{table}[t]
\centering
\caption{CIFAR-10 comparison under the WRN-34-10 backbone without data augmentation.}
\label{tb:wrn34_c10_no_aug}
\resizebox{\columnwidth}{!}{
{\arrayrulecolor{black}
\begin{tabular}{lccc}
\toprule
\textbf{Method} & \textbf{Clean (\%)} & \textbf{AA (\%)} & \textbf{Gap (\%)} \\
\midrule
MART~\citep{MART} & 84.17 & 51.10 & 9.90 \\
AWP~\citep{AWP} & 85.57 & 53.90 & 7.51 \\
FSR~\citep{FSR} & 84.46 & 53.03 & 7.99 \\
CFA~\citep{CFA} & 84.90 & 52.63 & 11.27 \\
UIAT~\citep{UIAT} & 83.20 & 52.92 & 9.40 \\
SGLR~\citep{SGLR} & 85.81 & 52.77 & 9.84 \\
DHAT~\citep{zhang2025towards} & 86.10 & 54.48 & 6.30 \\
\textbf{HICAT~(Ours)} & \textbf{87.44} & \textbf{56.20} & \textbf{3.03} \\
\bottomrule
\end{tabular}}}
\end{table}

\begin{table}[t]
\centering
\caption{CIFAR-10 semi-supervised comparison under the WRN-28-10 backbone with the RST 500K-unlabeled setting.}
\label{tb:wrn28_rst_c10}
\resizebox{\columnwidth}{!}{
{\arrayrulecolor{black}
\begin{tabular}{lccc}
\toprule
\textbf{Method} & \textbf{Clean (\%)} & \textbf{AA (\%)} & \textbf{Gap (\%)} \\
\midrule
MART~\citep{MART} & 84.70 & 52.30 & 8.40 \\
AWP~\citep{AWP} & 84.60 & 53.80 & 5.90 \\
FSR~\citep{FSR} & 84.40 & 53.70 & 6.75 \\
CFA~\citep{CFA} & 86.30 & 54.25 & 8.65 \\
UIAT~\citep{UIAT} & 87.25 & 55.20 & 9.00 \\
SGLR~\citep{SGLR} & 88.25 & 53.95 & 10.80 \\
DHAT~\citep{zhang2025towards} & 88.45 & 56.45 & 2.25 \\
\textbf{HICAT~(Ours)} & \textbf{89.10} & \textbf{57.55} & \textbf{1.95} \\
\bottomrule
\end{tabular}}}
\end{table}

\begin{table*}[t]
\centering
\caption{
Cross-model transfer attack accuracy~(\%) under the single-model transfer setting.
Adversarial examples are crafted on a source model and evaluated on a different target model.
We retain six representative transfer settings to fit the expanded baseline set.
The \textbf{bold} numbers indicate the best performance.
}
\label{tb:trans_attack}
\scriptsize
\setlength{\tabcolsep}{2.2pt}
\renewcommand{\arraystretch}{0.95}
\resizebox{0.96\textwidth}{!}{
{\arrayrulecolor{black}
\begin{tabular}{c|ccccc|ccccc|ccccc}
\toprule
\multirow{2}{*}{\textbf{Attack}}
& \multicolumn{5}{c|}{\textbf{VGG-16 $\rightarrow$ WRN28-10}}
& \multicolumn{5}{c|}{\textbf{VGG-16 $\rightarrow$ Inc-V3}}
& \multicolumn{5}{c}{\textbf{WRN28-10 $\rightarrow$ ResNet-50}} \\
\cmidrule(lr){2-6}\cmidrule(lr){7-11}\cmidrule(lr){12-16}
& MART & SGLR & DHAT & UIAT & \textbf{HICAT}
& MART & SGLR & DHAT & UIAT & \textbf{HICAT}
& MART & SGLR & DHAT & UIAT & \textbf{HICAT} \\
\midrule
FGSM & 61.93 & 64.03 & 64.55 & 63.73 & \textbf{65.36} & 61.64 & 63.74 & 64.21 & 63.44 & \textbf{64.98} & 75.31 & 77.41 & 78.02 & 77.11 & \textbf{78.92} \\
PGD-10 & 52.37 & 54.47 & 54.91 & 54.17 & \textbf{55.64} & 50.91 & 53.11 & 53.47 & 52.81 & \textbf{54.12} & 60.13 & 62.23 & 62.55 & 61.93 & \textbf{63.18} \\
PGD-20 & 52.00 & 54.10 & 54.64 & 53.80 & \textbf{55.47} & 50.83 & 52.93 & 53.32 & 52.63 & \textbf{54.01} & 59.72 & 61.82 & 62.23 & 61.52 & \textbf{62.94} \\
PGD-50 & 51.97 & 53.97 & 54.65 & 53.77 & \textbf{55.53} & 50.72 & 52.82 & 53.34 & 52.62 & \textbf{54.05} & 59.71 & 61.81 & 62.25 & 61.51 & \textbf{62.98} \\
C\&W & 53.51 & 55.61 & 55.99 & 55.31 & \textbf{56.67} & 49.58 & 51.68 & 52.14 & 51.38 & \textbf{52.89} & 59.25 & 61.35 & 61.73 & 61.05 & \textbf{62.41} \\
AA & 57.85 & 59.95 & 60.39 & 59.65 & \textbf{61.12} & 54.22 & 56.32 & 56.82 & 56.02 & \textbf{57.61} & 65.38 & 67.48 & 68.05 & 67.18 & \textbf{68.92} \\
\bottomrule
\end{tabular}}}

\vspace{0.35em}

\resizebox{0.96\textwidth}{!}{
{\arrayrulecolor{black}
\begin{tabular}{c|ccccc|ccccc|ccccc}
\toprule
\multirow{2}{*}{\textbf{Attack}}
& \multicolumn{5}{c|}{\textbf{WRN28-10 $\rightarrow$ VGG-16}}
& \multicolumn{5}{c|}{\textbf{ResNet-18 $\rightarrow$ WRN28-10}}
& \multicolumn{5}{c}{\textbf{ResNet-18 $\rightarrow$ Inc-V3}} \\
\cmidrule(lr){2-6}\cmidrule(lr){7-11}\cmidrule(lr){12-16}
& MART & SGLR & DHAT & UIAT & \textbf{HICAT}
& MART & SGLR & DHAT & UIAT & \textbf{HICAT}
& MART & SGLR & DHAT & UIAT & \textbf{HICAT} \\
\midrule
FGSM & 76.28 & 78.38 & 78.81 & 78.08 & \textbf{79.54} & 71.21 & 73.31 & 73.97 & 73.01 & \textbf{74.93} & 70.63 & 72.73 & 73.03 & 72.43 & \textbf{73.62} \\
PGD-10 & 62.93 & 64.03 & 65.33 & 64.73 & \textbf{65.93} & 56.76 & 58.86 & 59.59 & 58.56 & \textbf{60.62} & 56.02 & 58.12 & 58.55 & 57.82 & \textbf{59.27} \\
PGD-20 & 62.80 & 64.90 & 65.20 & 64.60 & \textbf{65.79} & 56.33 & 58.43 & 59.28 & 58.13 & \textbf{60.43} & 55.68 & 57.78 & 58.26 & 57.48 & \textbf{59.04} \\
PGD-50 & 62.76 & 64.86 & 65.22 & 64.56 & \textbf{65.88} & 56.35 & 58.45 & 59.35 & 58.15 & \textbf{60.55} & 55.65 & 57.75 & 58.32 & 57.45 & \textbf{59.18} \\
C\&W & 61.52 & 63.62 & 64.02 & 63.32 & \textbf{64.71} & 55.93 & 58.03 & 59.11 & 57.73 & \textbf{60.48} & 55.91 & 58.01 & 58.68 & 57.71 & \textbf{59.64} \\
AA & 70.06 & 72.16 & 72.65 & 71.86 & \textbf{73.44} & 60.56 & 62.66 & 63.10 & 62.36 & \textbf{63.84} & 58.89 & 60.99 & 61.53 & 60.69 & \textbf{62.37} \\
\bottomrule
\end{tabular}}}
\arrayrulecolor{black}
\end{table*}

\begin{table}[t]
\centering
\caption{
Cross-dataset transfer attack accuracy~(\%).
Adversarial examples are crafted from a source dataset and evaluated on a model trained on a different target dataset.
The \textbf{bold} numbers indicate the best performance.
}
\label{tb:cross_dataset_transfer}
\resizebox{0.98\columnwidth}{!}{
{\arrayrulecolor{black}
\begin{tabular}{c|c|ccccc}
\toprule
\textbf{Attack} & \textbf{Transfer} & MART & SGLR & DHAT & UIAT & \textbf{HICAT} \\
\midrule
\multicolumn{7}{c}{\textit{CIFAR Cross-Dataset Transfer ($\epsilon=8/255$)}} \\
\midrule
FGSM
& CIFAR-10 $\rightarrow$ CIFAR-100 & 60.41 & 62.69 & 63.07 & 62.10 & \textbf{63.55} \\
& CIFAR-100 $\rightarrow$ CIFAR-10 & 70.62 & 73.02 & 73.55 & 72.45 & \textbf{74.12} \\
\midrule
PGD-10
& CIFAR-10 $\rightarrow$ CIFAR-100 & 50.11 & 52.36 & 52.88 & 51.92 & \textbf{53.44} \\
& CIFAR-100 $\rightarrow$ CIFAR-10 & 58.49 & 60.91 & 61.43 & 60.36 & \textbf{62.10} \\
\midrule
PGD-20
& CIFAR-10 $\rightarrow$ CIFAR-100 & 49.70 & 51.86 & 52.41 & 51.43 & \textbf{52.98} \\
& CIFAR-100 $\rightarrow$ CIFAR-10 & 57.93 & 60.36 & 60.88 & 59.82 & \textbf{61.40} \\
\midrule
PGD-50
& CIFAR-10 $\rightarrow$ CIFAR-100 & 49.42 & 51.58 & 52.11 & 51.10 & \textbf{52.70} \\
& CIFAR-100 $\rightarrow$ CIFAR-10 & 57.56 & 59.96 & 60.48 & 59.40 & \textbf{61.05} \\
\midrule
C\&W
& CIFAR-10 $\rightarrow$ CIFAR-100 & 48.86 & 50.81 & 51.37 & 50.32 & \textbf{51.91} \\
& CIFAR-100 $\rightarrow$ CIFAR-10 & 56.12 & 58.51 & 58.97 & 57.95 & \textbf{59.38} \\
\midrule
AA
& CIFAR-10 $\rightarrow$ CIFAR-100 & 53.02 & 55.06 & 55.49 & 54.65 & \textbf{56.03} \\
& CIFAR-100 $\rightarrow$ CIFAR-10 & 61.41 & 63.55 & 64.08 & 63.10 & \textbf{64.72} \\
\midrule
\multicolumn{7}{c}{\textit{ImageNet Cross-Dataset Transfer ($\epsilon=4/255$)}} \\
\midrule
FGSM
& ImageNet-C $\rightarrow$ ImageNet-1K & 40.12 & 41.88 & 42.21 & 41.52 & \textbf{42.73} \\
PGD-10
& ImageNet-C $\rightarrow$ ImageNet-1K & 33.45 & 35.21 & 35.60 & 34.91 & \textbf{36.12} \\
PGD-20
& ImageNet-C $\rightarrow$ ImageNet-1K & 32.81 & 34.58 & 34.97 & 34.20 & \textbf{35.48} \\
PGD-50
& ImageNet-C $\rightarrow$ ImageNet-1K & 31.78 & 33.36 & 33.59 & 33.10 & \textbf{33.92} \\
C\&W
& ImageNet-C $\rightarrow$ ImageNet-1K & 30.54 & 32.29 & 32.62 & 31.85 & \textbf{33.10} \\
AA
& ImageNet-C $\rightarrow$ ImageNet-1K & 28.36 & 30.06 & 30.43 & 29.74 & \textbf{30.85} \\
\bottomrule
\end{tabular}
}
}
\arrayrulecolor{black}
\end{table}

\begin{table*}[t]
    \centering
    \caption{
    Performance of adversarial training methods under data augmentation.
    Models are trained on diffusion-generated samples and evaluated on CIFAR-10 and CIFAR-100.
    HICAT achieves the best clean accuracy, highest robustness, and lowest robust generalization gap.
    The \textbf{bold} numbers indicate the best performance.
    }
    \label{st:DDPM_data}
    \resizebox{\textwidth}{!}{
    \arrayrulecolor{black}
    \begin{tabular}{l|ccc|ccc|ccc|ccc}
    \toprule[1.5pt]
    & \multicolumn{6}{c|}{\textbf{CIFAR-10}}
    & \multicolumn{6}{c}{\textbf{CIFAR-100}} \\
    \cmidrule(lr){2-7} \cmidrule(lr){8-13}
    & \multicolumn{3}{c}{ResNet-18}
    & \multicolumn{3}{c|}{WRN28-10}
    & \multicolumn{3}{c}{ResNet-18}
    & \multicolumn{3}{c}{WRN28-10} \\
    \cmidrule(lr){2-4} \cmidrule(lr){5-7}
    \cmidrule(lr){8-10} \cmidrule(lr){11-13}
    \textbf{Method}
    & Clean$\uparrow$ & AA$\uparrow$ & Gap$\downarrow$
    & Clean$\uparrow$ & AA$\uparrow$ & Gap$\downarrow$
    & Clean$\uparrow$ & AA$\uparrow$ & Gap$\downarrow$
    & Clean$\uparrow$ & AA$\uparrow$ & Gap$\downarrow$ \\
    \midrule

    \multirow{1}{*}{MART~\citep{MART}}
    & 83.45 & 49.45 & 2.91
    & 84.26 & 51.95 & 8.73
    & 54.73 & 27.70 & 2.88
    & 55.87 & 30.20 & 8.86 \\

    \multirow{1}{*}{AWP~\citep{AWP}}
    & 83.78 & 50.79 & 2.03
    & 84.10 & 53.29 & 6.11
    & 57.55 & 29.33 & 2.37
    & 59.71 & 31.83 & 7.13 \\

    \multirow{1}{*}{FSR~\citep{FSR}}
    & 83.19 & 50.53 & 2.35
    & 83.88 & 53.03 & 7.05
    & 58.10 & 28.94 & 2.47
    & 59.03 & 30.44 & 7.40 \\

    \multirow{1}{*}{CFA~\citep{CFA}}
    & 84.97 & 50.85 & 2.98
    & 85.81 & 53.35 & 8.94
    & 60.13 & 28.85 & 3.01
    & 61.56 & 29.61 & 9.13 \\

    \multirow{1}{*}{UIAT~\citep{UIAT}}
    & 85.10 & 52.09 & 3.04
    & 86.73 & 54.59 & 9.41
    & 59.92 & 28.48 & 4.47
    & 60.24 & 30.98 & 14.32 \\

    \multirow{1}{*}{SGLR~\citep{SGLR}}
    & 86.35 & 51.27 & 2.75
    & 87.72 & 53.77 & 11.25
    & 61.25 & 29.10 & 4.15
    & 62.39 & 30.15 & 17.05 \\

    \multirow{1}{*}{DHAT~\citep{zhang2025towards}}
    & 87.62 & 54.42 & 0.83
    & 88.94 & 56.92 & 2.48
    & 63.14 & 32.21 & 0.98
    & 64.80 & 34.71 & 2.96 \\

    \multirow{1}{*}{\textbf{HICAT (Ours)}}
    & \textbf{88.40} & \textbf{55.63} & \textbf{0.71}
    & \textbf{89.85} & \textbf{58.11} & \textbf{2.12}
    & \textbf{64.52} & \textbf{33.48} & \textbf{0.86}
    & \textbf{66.02} & \textbf{36.20} & \textbf{2.51} \\

    \bottomrule[1.5pt]
    \end{tabular}
    }
\end{table*}

\subsection{Cross-Architecture Robustness Generalization}
To examine generality, we evaluate HICAT on convolutional networks (ResNet-50, VGG-16, Inception-V3) on CIFAR-10 (Table~\ref{st:different_net}) and on ViT-B/16 on ImageNet (Table~\ref{tb:imagenet_vit_b16}).
We also report two additional CIFAR-10 settings requested by the reviewers: WRN-34-10 without data augmentation (Table~\ref{tb:wrn34_c10_no_aug}) and WRN-28-10 with RST 500K unlabeled data~\citep{carmon2019unlabeled} (Table~\ref{tb:wrn28_rst_c10}). In the WRN-34-10 setting, HICAT improves AA over DHAT from 54.48\% to 56.20\% and reduces the Gap from 6.30\% to 3.03\%. Under the RST setting, HICAT improves AA from 56.45\% to 57.55\% and further reduces the Gap from 2.25\% to 1.95\%. These results show that the gains remain under both standard WRN training and semi-supervised training with unlabeled data.

\begin{itemize}
    \item \textbf{Consistent robustness gains across diverse extractors.}
    HICAT improves AutoAttack robustness on every tested architecture, outperforming DHAT by +1.32\% on ResNet-50 and +1.33\% on ImageNet with ViT-B/16.

    \item \textbf{No sacrifice in clean accuracy via precise diagnosis.}
    HICAT also improves clean accuracy over DHAT by +1.68\% on ResNet-50 and +1.18\% on ImageNet with ViT-B/16.

    \item \textbf{Reduced robust generalization gap via latent regularization.}
    HICAT achieves the smallest robust generalization gap across all tested models, including 1.72\% on ResNet-50 and 9.79\% on ImageNet with ViT-B/16.
\end{itemize}

\subsection{Transfer Attacks (Black-Box Robustness)}
To evaluate black-box robustness, we examine HICAT's resistance to adversarial examples transferred across \emph{different architectures} and \emph{different datasets} (Tables~\ref{tb:trans_attack} and \ref{tb:cross_dataset_transfer}). Table~\ref{tb:trans_attack} reports cross-model transfer under the expanded baseline set using six representative source--target pairs. Table~\ref{tb:cross_dataset_transfer} reports the corresponding multi-baseline cross-dataset transfer comparison.

\subsubsection{Cross-Model Transferability}
HICAT consistently outperforms UIAT when facing attacks from heterogeneous architectures.
\begin{itemize}
    \item \textbf{Higher accuracy across model pairs.}
    HICAT improves AA accuracy by +1.47\% over UIAT (VGG-16 $\to$ WRN28-10) and +1.68\% (ResNet-18 $\to$ Inc-V3).
    Transfer attacks often exploit shared non-robust cues, including background textures. The gains suggest that FLOE reduces reliance on these cues, making transferred perturbations less effective.

    \item \textbf{Larger improvements under stronger attacks.}
    Across the selected transfer pairs, HICAT consistently improves over UIAT under PGD-50 and AA.
    Strong attacks tend to overfit the gradient landscape of the source model. Adaptive Debiasing aligns predictions with causal semantics rather than noise, making source-model attack directions less effective.
\end{itemize}

\subsubsection{Cross-Dataset Transfer}
We further test generalization across datasets with different label distributions.
\begin{itemize}
    \item \textbf{Stable improvements across CIFAR datasets.}
    With MART, SGLR, DHAT, and UIAT jointly compared, HICAT achieves the highest accuracy across all attacks in both transfer directions (CIFAR-10 $\Leftrightarrow$ CIFAR-100). Under AA, HICAT improves over DHAT by +0.54\% (CIFAR-10 $\rightarrow$ CIFAR-100) and +0.64\% (CIFAR-100 $\rightarrow$ CIFAR-10).
    These results indicate that separating foreground semantics from environmental noise remains effective when the label space changes. By reducing reliance on low-level dataset artifacts, HICAT improves transfer robustness across CIFAR datasets.

    \item \textbf{ImageNet-C transfer.}
    Under corruption-induced perturbations, HICAT improves AutoAttack accuracy by +0.42\% over DHAT and +0.79\% over SGLR.
    Common corruptions such as snow often appear as image-wide background artifacts.
    LBBE assigns low causal relevance to these artifacts, reducing their influence on semantic feature extraction. This suggests that causal alignment also helps under corruption-induced distribution shifts.
\end{itemize}

\subsection{Compatibility with Data Augmentation}

We evaluate HICAT's synergy with synthetic data by augmenting CIFAR-10 and CIFAR-100 with 50K DDPM samples~\citep{wang2023better} under a unified protocol (Table~\ref{st:DDPM_data}).

\begin{itemize}
    \item \textbf{Significant clean accuracy improvements.}
    HICAT achieves the highest clean accuracy, surpassing UIAT by +3.12\% on CIFAR-10 (WRN28-10) to reach 89.85\%.
    This suggests that LBBE helps retain clean accuracy when generated samples are added, rather than improving robustness only by sacrificing standard accuracy.

    \item \textbf{Consistent boosts in robust accuracy.}
    HICAT yields consistent AA robustness gains of 3.5--5.3\% over UIAT. On CIFAR-100 with WRN28-10, AA accuracy improves from 30.98\% (UIAT) to 36.20\% (+5.22\%).
    These gains indicate that Adaptive Debiasing remains effective under the enlarged training distribution and improves robust accuracy under the same augmentation budget.

    \item \textbf{Substantial reduction in robust generalization gap.}
    HICAT minimizes the robust generalization gap, reducing it from 14.32\% (UIAT) to 2.51\% on CIFAR-100 (WRN28-10).
    This result indicates that FLOE reduces robust overfitting in the augmented setting, as reflected by the lower train--test robust gap.
\end{itemize}

\begin{table}[t]
    \centering
    \providecommand{\LargeScaleAugTableWidth}{\columnwidth}
    \caption{Comparison of HICAT with large-scale data-augmentation baselines on CIFAR-10 under AutoAttack ($\epsilon=8/255$). HICAT rows report a data-scaling sweep with a fixed 100-epoch schedule; at 200K generated samples, HICAT obtains AA close to the reported Gowal et al.\ result under this evaluation setting.}
    \label{tb:large_scale_aug}
    \resizebox{\LargeScaleAugTableWidth}{!}{
    {\arrayrulecolor{black}
    \begin{tabular}{lccccc}
        \toprule
        \textbf{Method} & \textbf{Backbone} & \textbf{Aug.\ Data} & \textbf{Epochs} & \textbf{Clean (\%)} & \textbf{AA (\%)} \\
        \midrule
        Gowal et al.~\citep{gowal2021improving} & WRN-70-16 & 100M & 2000 & 88.94 & 62.20 \\
        Gowal et al.~\citep{gowal2021improving} & WRN-28-10 & 100M & 2000 & 87.37 & 61.38 \\
        Pang et al.~\citep{pang2022robustness}  & WRN-70-16 & 1M   & 800  & 88.70 & 60.20 \\
        Pang et al.~\citep{pang2022robustness}  & WRN-28-10 & 1M   & 800  & 88.02 & 57.52 \\
        \midrule
        HICAT (Ours) & WRN-70-16 & 5K   & 100 & 89.12 & 57.83 \\
        HICAT (Ours) & WRN-70-16 & 50K  & 100 & 90.44 & 60.81 \\
        HICAT (Ours) & WRN-70-16 & 100K & 100 & 90.75 & 61.68 \\
        HICAT (Ours) & WRN-70-16 & 200K & 100 & 91.08 & 62.17 \\
        \midrule
        HICAT (Ours) & WRN-28-10 & 5K   & 100 & 86.74 & 55.38 \\
        HICAT (Ours) & WRN-28-10 & 50K  & 100 & 89.85 & 58.11 \\
        HICAT (Ours) & WRN-28-10 & 100K & 100 & 90.18 & 59.96 \\
        HICAT (Ours) & WRN-28-10 & 200K & 100 & 90.61 & 61.32 \\
        \bottomrule
    \end{tabular}}}
\end{table}

We further compare HICAT with large-scale augmentation baselines (Gowal et al.~\citep{gowal2021improving}: 100M DDPM samples and 2000 epochs; Pang et al.~\citep{pang2022robustness}: 1M DDPM samples and 800 epochs) under the same AutoAttack evaluation protocol on CIFAR-10 (Table~\ref{tb:large_scale_aug}). HICAT uses 5K--200K DDPM samples with a fixed 100-epoch schedule. We use 200K as the largest setting to keep the augmentation budget moderate and evaluate HICAT without using 1M--100M generated samples. Compared with Gowal et al., HICAT has higher Clean accuracy (+2.14\% on WRN-70-16 and +3.24\% on WRN-28-10), while AA is lower by only 0.03\% and 0.06\%, respectively.

We also evaluate HICAT under unlabeled-data supervision on CIFAR-10 (ResNet-18) with 500K unlabeled images following~\citep{carmon2019unlabeled}. MART, AWP, FSR, CFA, UIAT, SGLR, DHAT, and HICAT are compared under the same schedule and attack/evaluation protocol. As shown in Table~\ref{tb:semi_unlabeled_c10}, HICAT remains the top-performing method and improves over UIAT (Clean: +3.18\%, AA: +3.56\%, Gap: -2.36\%) and over DHAT in AA robustness (+1.51\%).

\begin{table}[t]
    \centering
    \providecommand{\SemiUnlabeledTableWidth}{\columnwidth}
    \caption{Semi-supervised results on CIFAR-10 (ResNet-18) with 500K unlabeled images under the unified evaluation protocol.}
    \label{tb:semi_unlabeled_c10}
    \resizebox{\SemiUnlabeledTableWidth}{!}{
    {\arrayrulecolor{black}
    \begin{tabular}{lccc}
        \toprule
        \textbf{Method} & \textbf{Clean (\%)} & \textbf{AA (\%)} & \textbf{Gap (\%)} \\
        \midrule
        MART~\citep{MART} & 82.91 & 48.86 & 3.20 \\
        AWP~\citep{AWP} & 83.26 & 50.11 & 2.36 \\
        FSR~\citep{FSR} & 82.78 & 49.84 & 2.63 \\
        CFA~\citep{CFA} & 84.31 & 50.22 & 3.31 \\
        UIAT~\citep{UIAT} & 84.78 & 51.62 & 3.32 \\
        SGLR~\citep{SGLR} & 85.67 & 50.93 & 3.08 \\
        DHAT~\citep{zhang2025towards} & 86.94 & 53.67 & 1.21 \\
        \textbf{HICAT (Ours)} & \textbf{87.96} & \textbf{55.18} & \textbf{0.96} \\
        \bottomrule
    \end{tabular}}}
\end{table}

\subsection{Robustness Under Varying Attack Strengths}

\begin{figure*}[t]
    \centering
    \setlength{\tabcolsep}{1pt}
    \renewcommand{\arraystretch}{1}

    \begin{tabular}{@{}ccccc@{}}
        \includegraphics[width=0.19\linewidth]{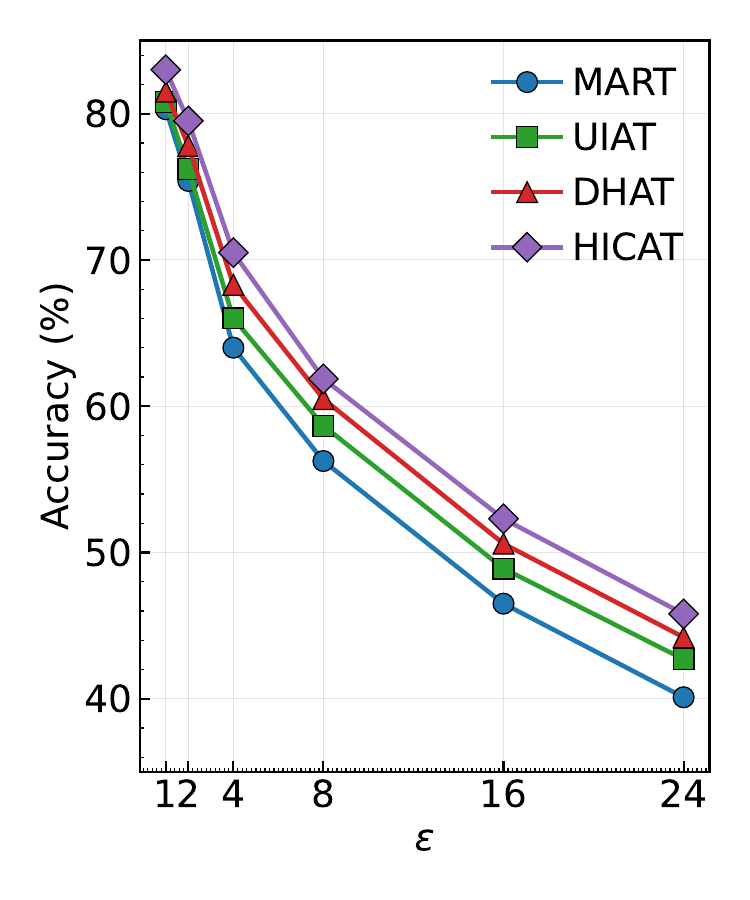} &
        \includegraphics[width=0.19\linewidth]{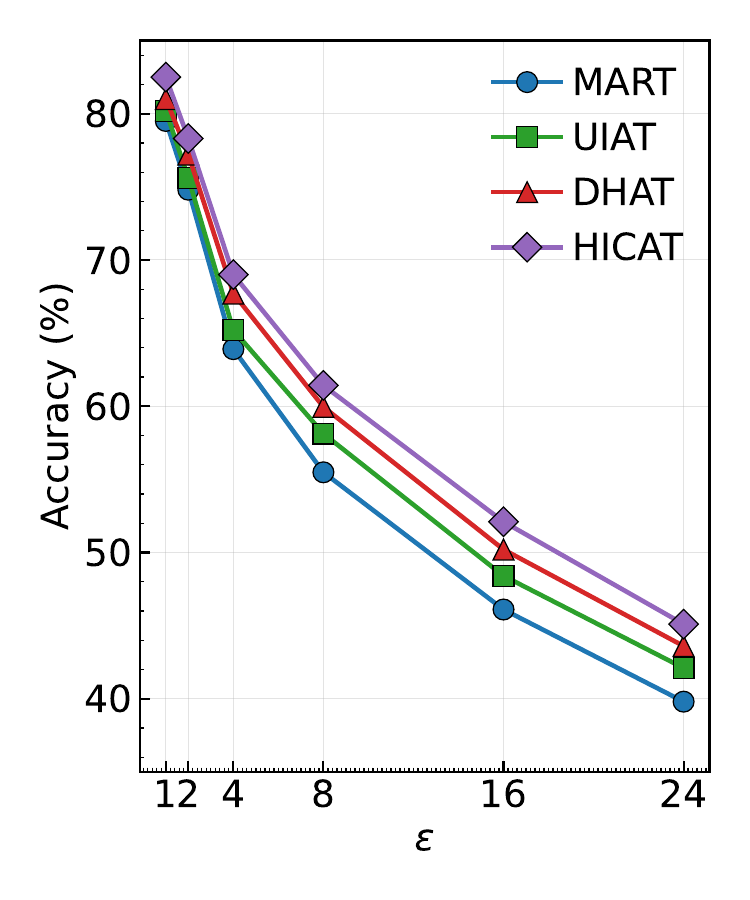} &
        \includegraphics[width=0.19\linewidth]{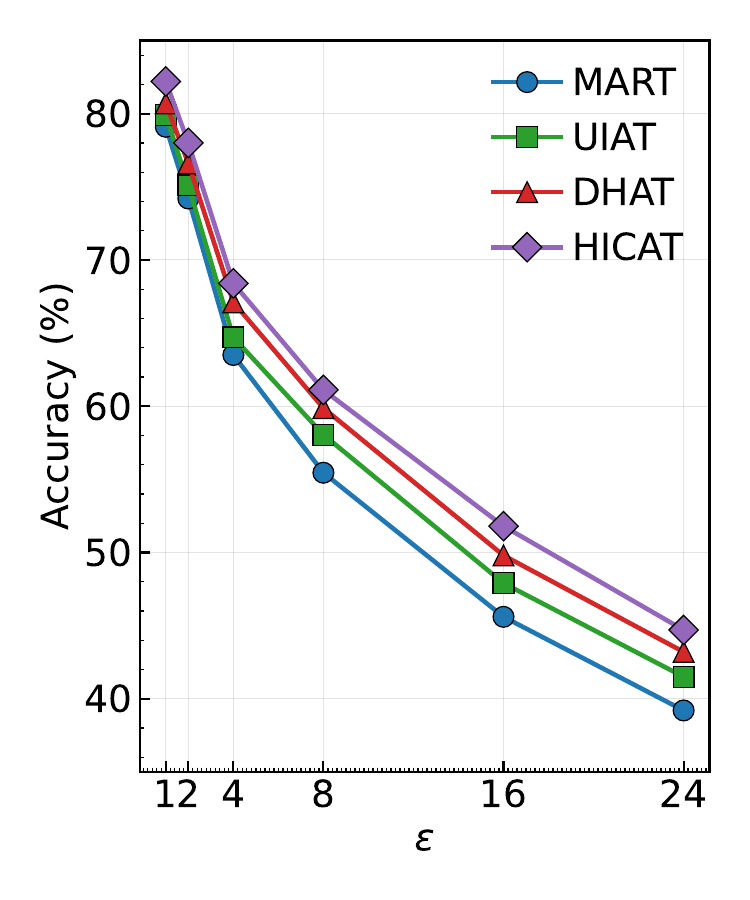} &
        \includegraphics[width=0.19\linewidth]{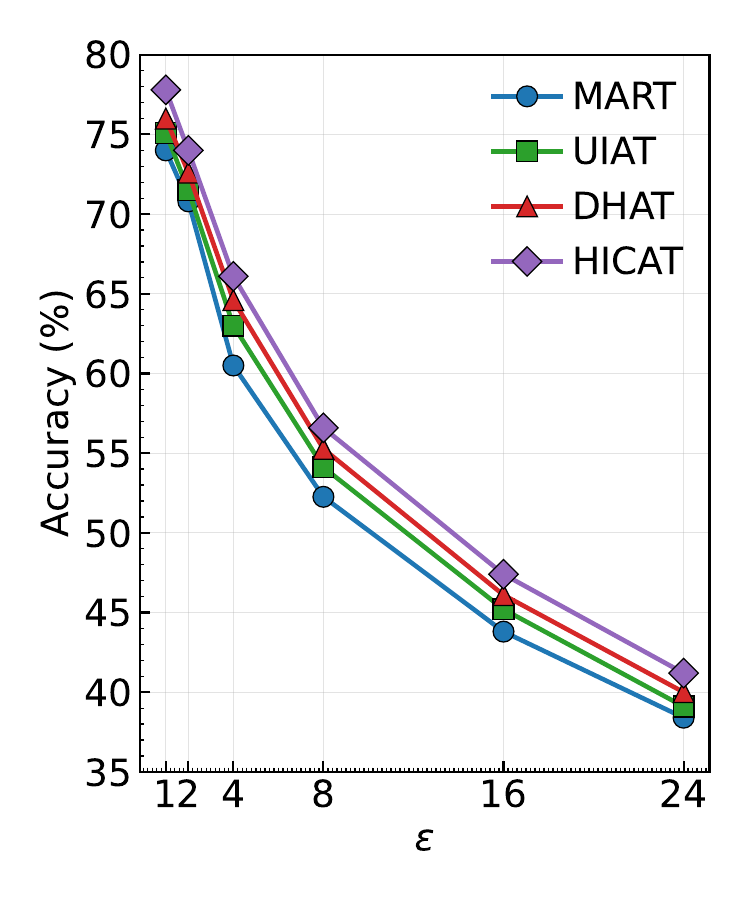} &
        \includegraphics[width=0.19\linewidth]{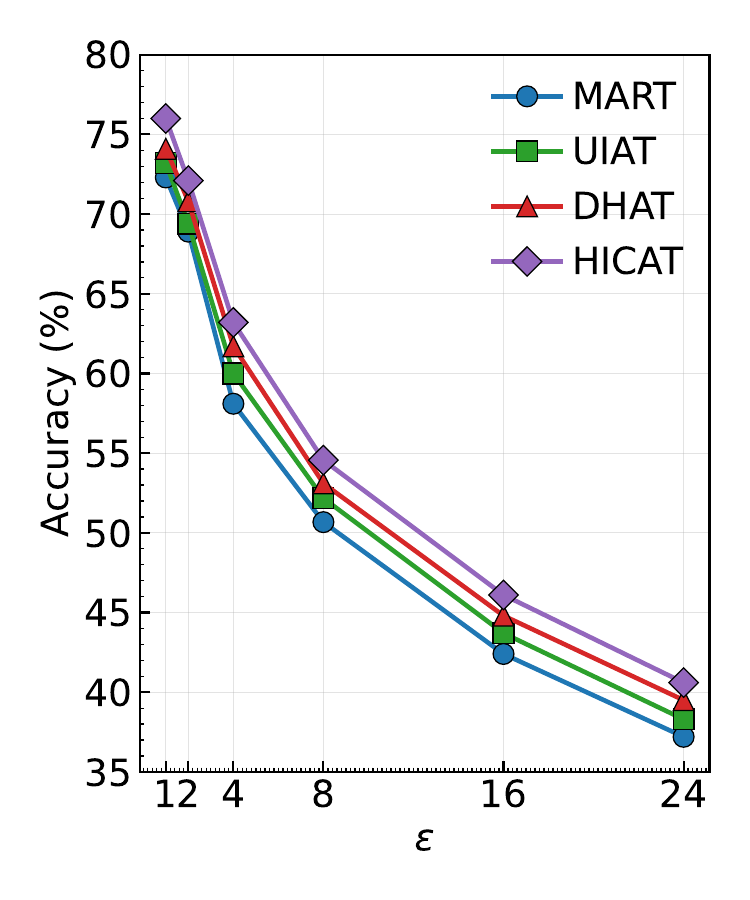} \\
        \footnotesize{(a) PGD-10 ($\epsilon$ sweep)} &
        \footnotesize{(b) PGD-20 ($\epsilon$ sweep)} &
        \footnotesize{(c) PGD-50 ($\epsilon$ sweep)} &
        \footnotesize{(d) C\&W ($\epsilon$ sweep)} &
        \footnotesize{(e) AA ($\epsilon$ sweep)}
    \end{tabular}

    \vspace{4pt}

    \begin{tabular}{@{}ccccc@{}}
        \includegraphics[width=0.19\linewidth]{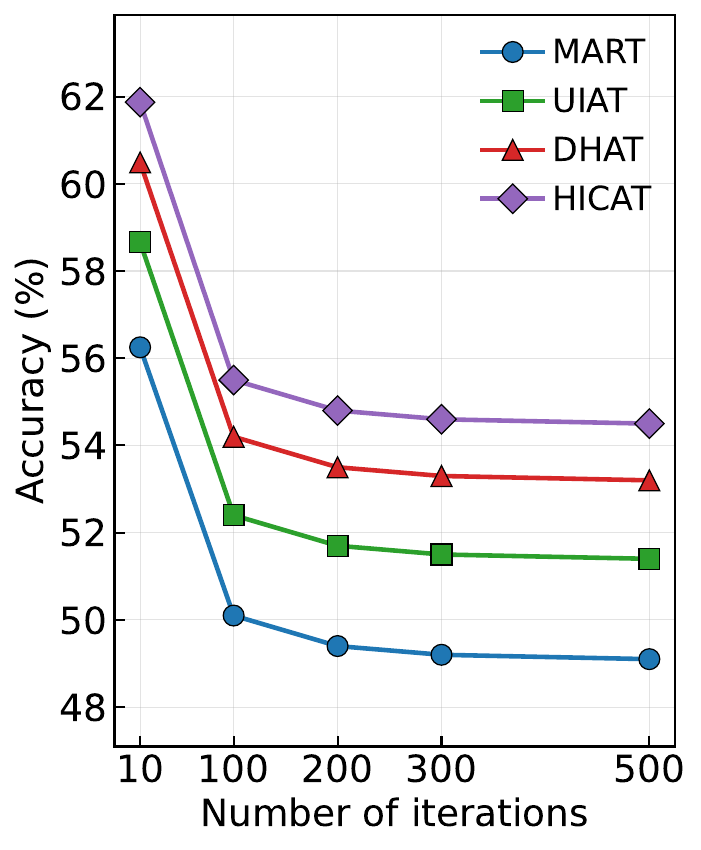} &
        \includegraphics[width=0.19\linewidth]{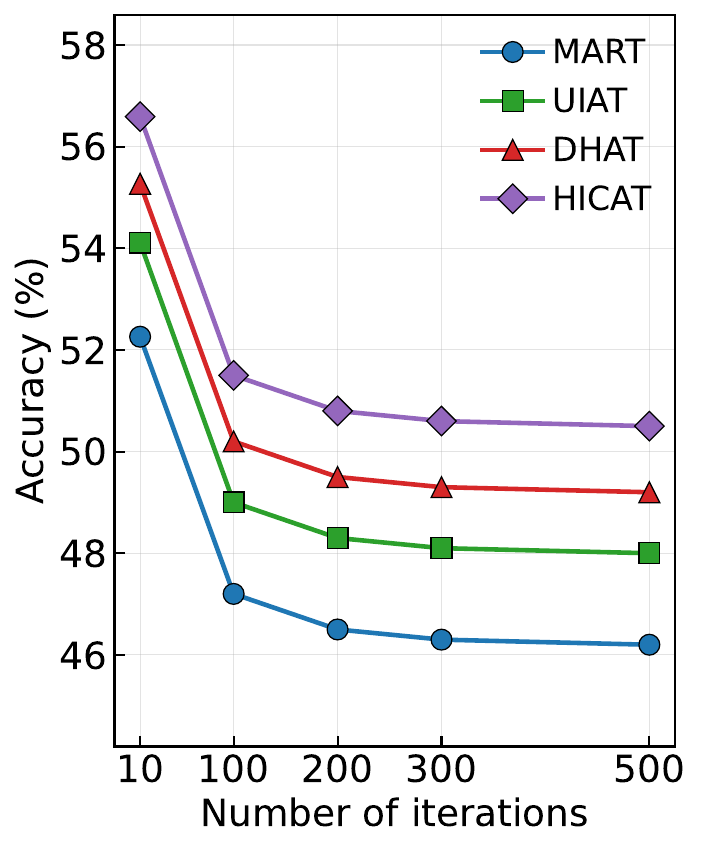} &
        \includegraphics[width=0.19\linewidth]{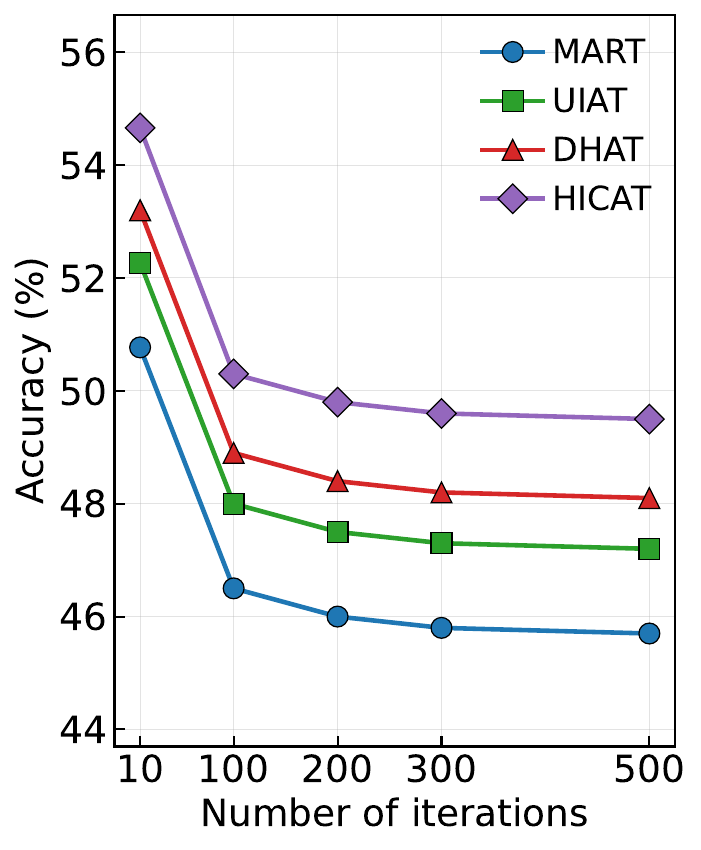} &
        \includegraphics[width=0.19\linewidth]{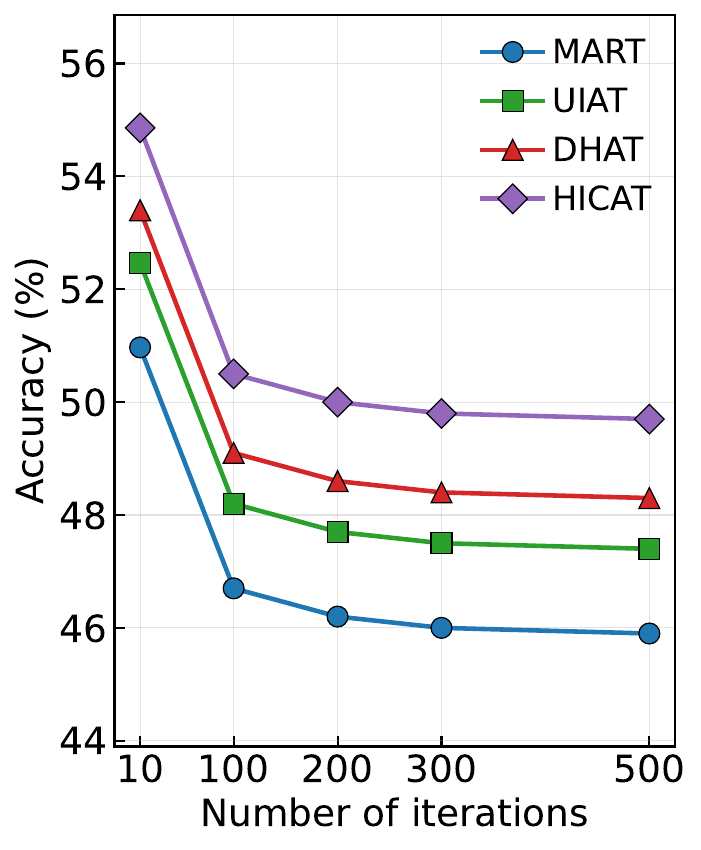} &
        \includegraphics[width=0.19\linewidth]{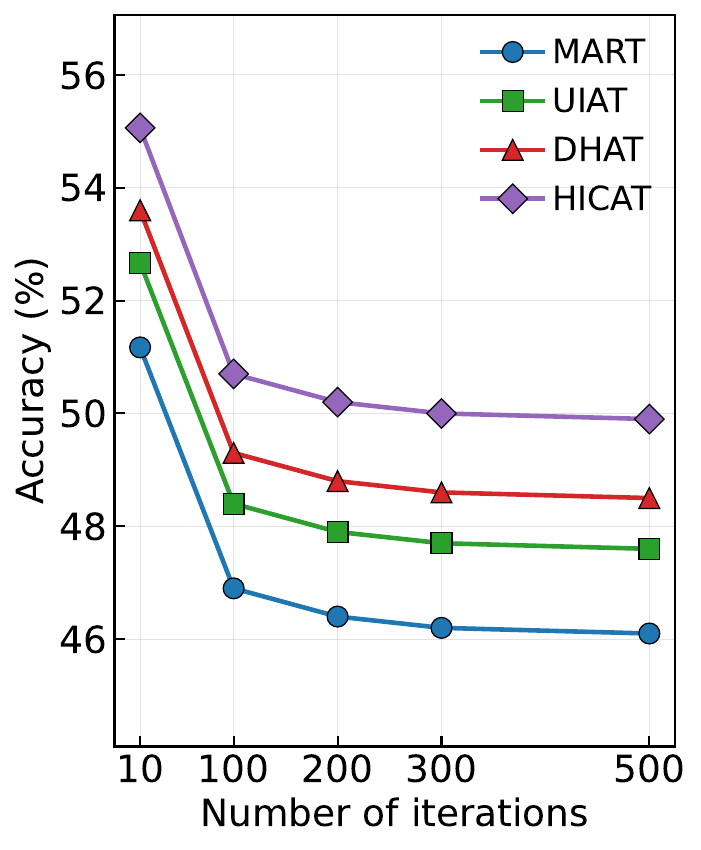} \\
        \footnotesize{(f) PGD (iter.)} &
        \footnotesize{(g) C\&W (iter.)} &
        \footnotesize{(h) APGD-CE (iter.)} &
        \footnotesize{(i) APGD-DLR (iter.)} &
        \footnotesize{(j) FAB (iter.)}
    \end{tabular}

    \caption{
        \textbf{Robustness evaluation under varying attack strengths on CIFAR-10 using WRN28-10.}
        We compare the defense performance of HICAT against strong matched baselines (MART, UIAT, DHAT) across diverse adversarial constraints.
        \textbf{Top Row (a--e):} Robustness under varying perturbation budgets $\epsilon \in \{1, 2, 4, 8, 16, 24\}/255$. HICAT consistently exhibits the highest accuracy, demonstrating substantial resilience even against large-magnitude perturbations.
        \textbf{Bottom Row (f--j):} Robustness under varying attack iterations (optimization steps) from 10 to 500, with fixed $\epsilon=8/255$. We examine PGD, C\&W, and the individual components of AutoAttack (APGD-CE, APGD-DLR, and FAB).
    }
    \label{fig:attack_strength_sweep}
\end{figure*}

To understand defense behavior under challenging settings, we analyze robustness from two perspectives: varying the perturbation budget and varying the optimization intensity. These experiments provide a fine-grained evaluation beyond fixed-budget benchmarks (Fig.~\ref{fig:attack_strength_sweep}).

\subsubsection{Varying Perturbation Budget}

We vary the perturbation magnitude $\epsilon \in \{1,2,4,8, \allowbreak 16,24\}/255$ under PGD, C\&W, and AutoAttack to examine degradation patterns.

\begin{itemize}
    \item \textbf{Highest robustness across all $\epsilon$ settings.}
    Even in the extreme regime ($\epsilon = 24/255$), HICAT maintains a noticeable margin over MART, UIAT, and DHAT, demonstrating superior resistance to large-magnitude perturbations.

    \item \textbf{Slower degradation with increasing $\epsilon$.}
    Unlike baselines that drop sharply, HICAT exhibits smooth decay. This confirms that Adaptive Debiasing effectively filters brittle background noise; by reducing reliance on high-frequency correlations, it forces attackers to target robust semantic features, thereby delaying performance collapse.

    \item \textbf{Stable behavior under AutoAttack.}
    While AA integrates multiple strong sub-attacks, HICAT remains consistently more robust, showing that the improvements extend beyond standard first-order gradient-based PGD variants.
\end{itemize}

\begin{figure*}[t]
    \centering
    \includegraphics[width=1.0\linewidth]{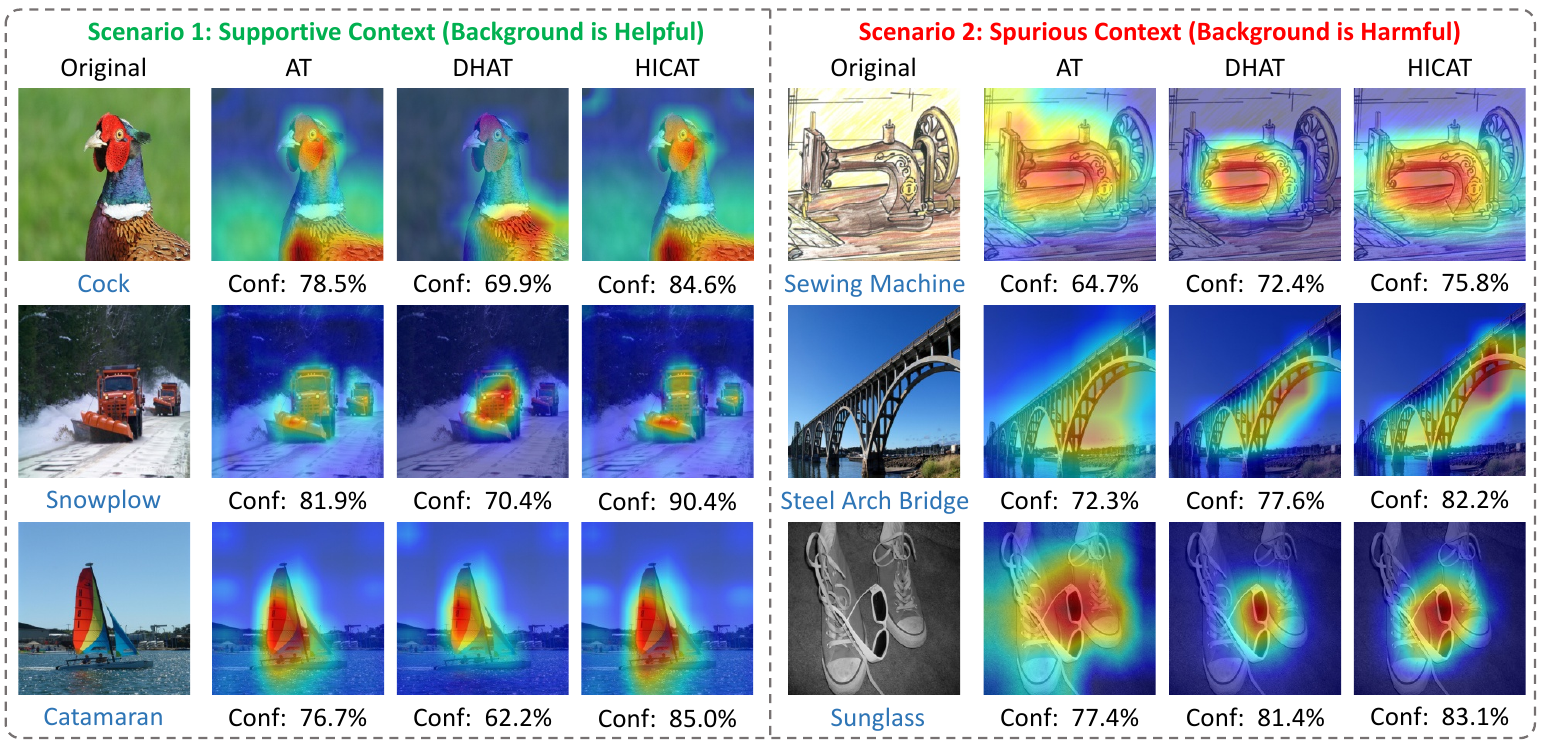}
    \caption{
    Visualization of feature activation maps using Standard AT, DHAT, and HICAT via Grad-CAM.
    The figure is divided into two scenarios: Supportive Context (Left Block) and Spurious Context (Right Block).
    Left-to-right in each block: the original image; feature activation maps from Standard AT/DHAT/HICAT(Ours).
    The prediction confidence for the ground-truth class is displayed below each visualization.
    }
    \label{fig:gradcam_vis}
\end{figure*}

\subsubsection{Varying Attack Steps}

To probe ``worst-case'' robustness, we increase attack iterations $K \in \{10, \dots, 500\}$ to allow adversaries ample budget to overcome gradient masking.

\begin{itemize}
    \item \textbf{Resistance to Gradient Masking.}
    HICAT maintains the highest accuracy throughout the steep optimization phase. Unlike MART and UIAT, which suffer sharp degradation as gradients become more informative, HICAT's stability suggests that it reduces reliance on non-robust features rather than only creating a loss landscape that temporarily obstructs optimization.

    \item \textbf{Superior Asymptotic Stability.}
    After 200 steps, HICAT remains higher than the baselines. This suggests that FLOE moves the decision boundary toward foreground semantics and leaves fewer background directions for the adversary.

    \item \textbf{Universality Across Optimization Landscapes.}
    HICAT remains stable against Cross-Entropy, Logit-Difference, and Projection-based objectives. This suggests that causal logit alignment improves robustness across different attack losses.
\end{itemize}

\subsection{Visual Explanations and Analysis}
\label{sec:visual_analysis}

To verify HICAT's Causal Alignment, we visualize Grad-CAM attention maps~\citep{grad_cam_paper} and prediction confidence in Figure~\ref{fig:gradcam_vis}. Comparing these behaviors highlights three advantages:
\begin{itemize}
    \item \textbf{Supportive Priors.}
    For context-dependent classes (\eg, \textit{Snowplow}), DHAT treats valid priors as noise due to rigid suppression, dropping confidence to 70.4\%. In contrast, HICAT identifies the background as supportive via LBBE and relaxes constraints. This preserves useful context and achieves the highest confidence (90.4\%) in this example.

    \item \textbf{Targeted Debiasing.}
    In object-centric cases (\eg, \textit{Sewing Machine}), Standard AT overfits to irrelevant textures, reducing confidence to 64.7\%. HICAT filters spurious backgrounds and sharpens focus on causal foreground structures. This targeted debiasing improves confidence over the baseline (\eg, {75.8\%} for \textit{Sewing Machine}, 83.1\% for \textit{Sunglasses}).

    \item \textbf{Dynamic Modulation.}
    The visual comparison shows that HICAT avoids the static behavior of prior baselines. It suppresses background features when they are spurious and preserves them when they aid recognition.
\end{itemize}

\begin{figure*}[t]
    \centering
    \setlength{\tabcolsep}{1pt}
    \renewcommand{\arraystretch}{1}
    \begin{tabular}{@{}ccc@{}}
        \includegraphics[width=0.33\linewidth]{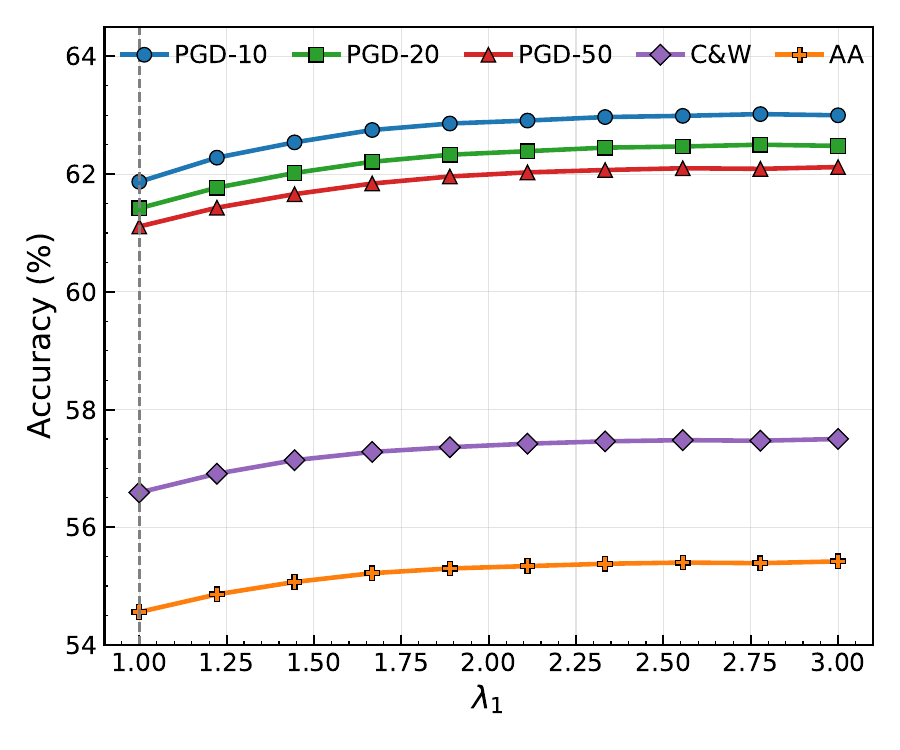} &
        \includegraphics[width=0.33\linewidth]{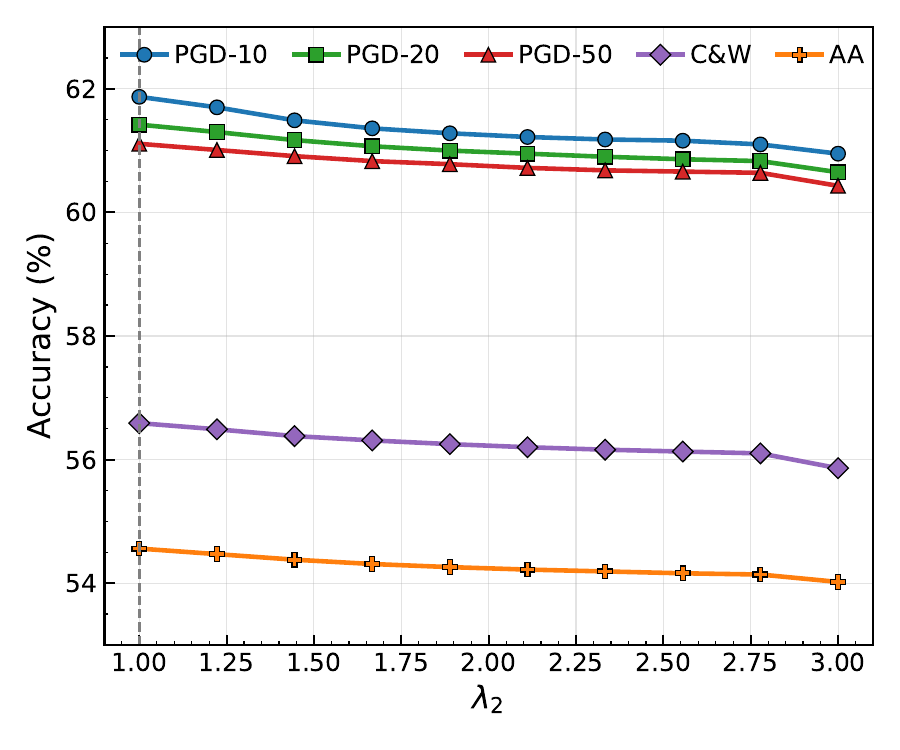} &
        \includegraphics[width=0.33\linewidth]{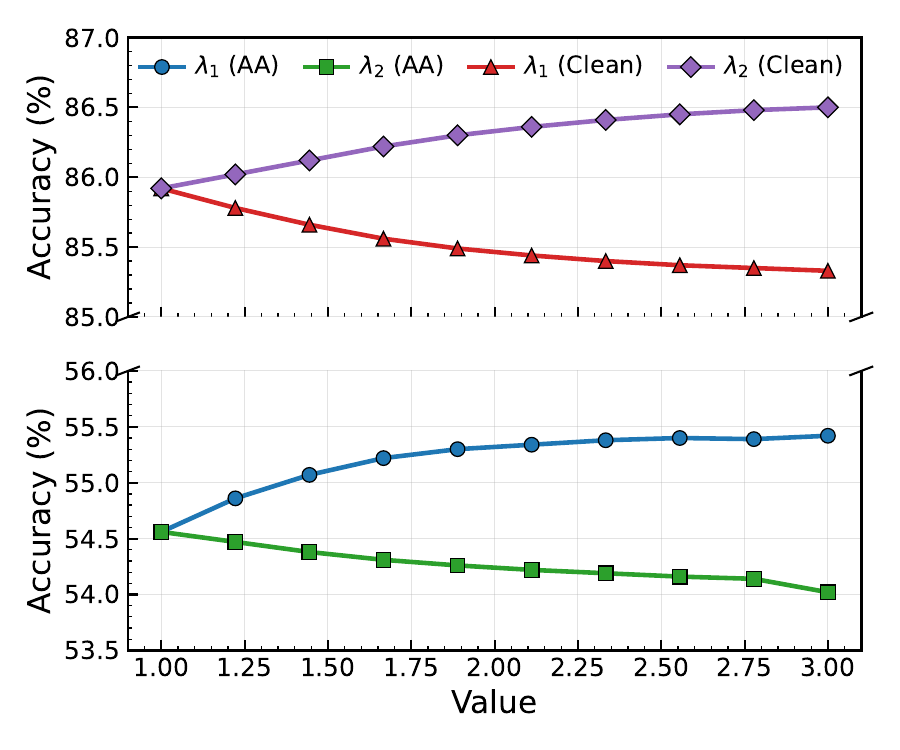} \\
        \footnotesize{(a) $\lambda_1$ sensitivity} &
        \footnotesize{(b) $\lambda_2$ sensitivity} &
        \footnotesize{(c) Clean\-/\-/Robust trade\-/off w.r.t. $\lambda$}
        \\[8pt]
        \includegraphics[width=0.33\linewidth]{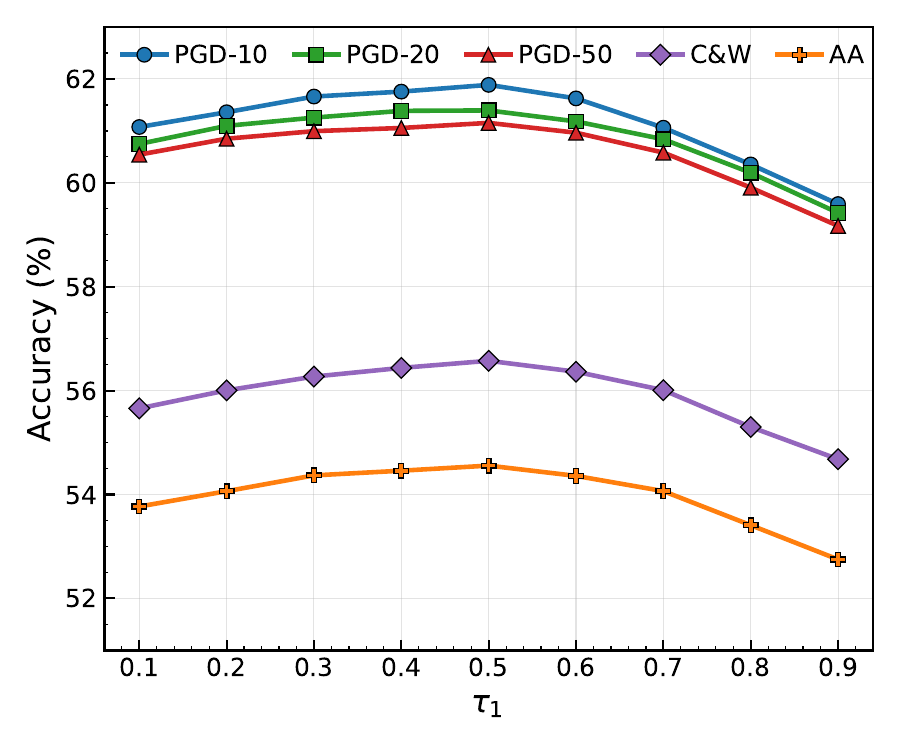} &
        \includegraphics[width=0.33\linewidth]{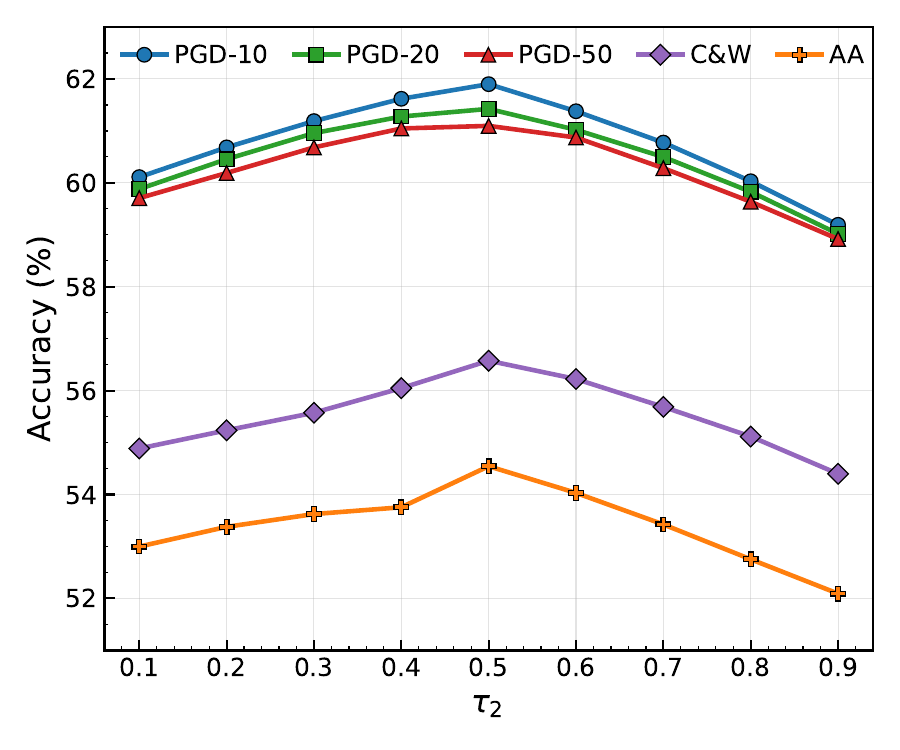} &
        \includegraphics[width=0.33\linewidth]{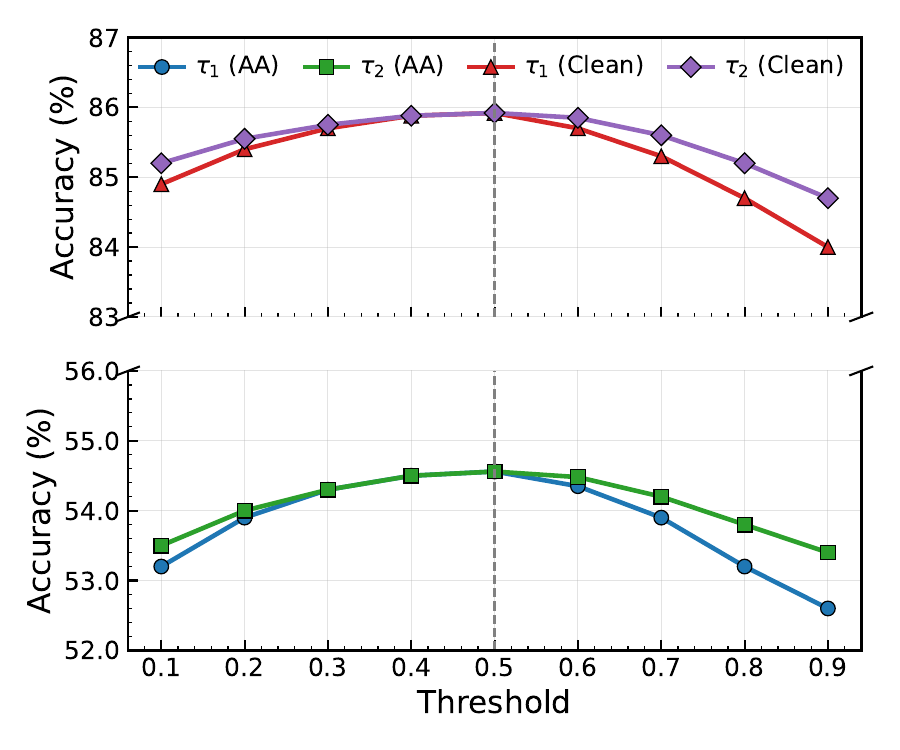} \\
        \footnotesize{(d) $\tau_1$ sensitivity} &
        \footnotesize{(e) $\tau_2$ sensitivity} &
        \footnotesize{(f) Clean\-/\-/Robust trade\-/off w.r.t. $\tau$}
    \end{tabular}
    \caption{
        \textbf{Parameter sensitivity analysis.}
        We evaluate the robustness of our method under different training and inference hyperparameters.
        (a,b) show the influence of logit debiasing weights $\lambda_1$ and $\lambda_2$ on robustness across different adversarial attacks.
        (c) visualizes the clean–robust accuracy trade\-/off under varying $\lambda$ values using a broken $y$\-/axis for clarity.
        (d,e) analyze the effect of foreground/background thresholding parameters $\tau_1$ and $\tau_2$, which control causal mask sparsity and background separation.
        (f) summarizes the corresponding clean–robust trade\-/off under different $\tau$ values.
        Across all settings, the curves demonstrate that HICAT remains stable within a broad hyperparameter range, achieving best performance around $\lambda = 1.0$ and $\tau = 0.5$.
    }
    \label{fig:parameter}
\end{figure*}

\subsection{Hyperparameter Sensitivity Analysis}

We conduct a comprehensive sensitivity study to examine the stability of HICAT under different training and inference hyperparameters. Figure~\ref{fig:parameter} summarizes the effects of the debiasing weights $(\lambda_1,\lambda_2)$ and thresholds $(\tau_1,\tau_2)$.
Three consistent observations emerge:

\begin{itemize}
    \item \textbf{Stable robustness across a wide range of $\lambda$ values.}
    Robustness improves as $\lambda_1$ or $\lambda_2$ increase to $1.0$ and then plateaus (Figs.~\ref{fig:parameter}(a)-(b)).
    This stability corroborates the \textit{structural nature} of our causal alignment. Unlike standard regularization where hyperparameters strictly balance loss magnitudes, HICAT alters the \textit{optimization direction} to enforce feature disentanglement. Once the representation is sufficiently orthogonalized (around $\lambda \approx 1.0$), increasing weights yields diminishing returns, confirming that robustness stems from learned feature invariance rather than gradient masking.

    \item \textbf{Complementary mechanisms of $\lambda_1$ and $\lambda_2$.}
    Figure~\ref{fig:parameter}(c) reveals that $\lambda_1$ and $\lambda_2$ regulate the trade-off through distinct pathways.
    $\lambda_1$ (Adaptive Debiasing) operates in the \textit{logit space}, purifying the supervision signal by filtering high-confidence noise.
    $\lambda_2$ (FLOE) operates in the \textit{geometric space}, enforcing structural orthogonality. While stronger $\lambda_2$ favors clean feature separation, it imposes a rigid constraint that slightly limits flexibility for adversarial adaptation.
    The results confirm that $\lambda_1=\lambda_2=1.0$ achieves optimal synergy between signal purification and geometric disentanglement.

    \item \textbf{Foreground/background thresholds $(\tau_1,\tau_2)$ exhibit smooth optima.}
    Robustness peaks around $\tau_1=\tau_2=0.5$ (Figs.~\ref{fig:parameter}(d)-(e)).
    This reflects the \textit{semantic signal-to-noise ratio} in mask generation.
    Lower thresholds ($\tau < 0.3$) fail to isolate the foreground, leaking spurious background noise into the causal estimator. Conversely, overly high thresholds ($\tau > 0.7$) aggressively crop the object, causing the loss of essential semantic features.
    The peak at $0.5$ represents a Causal Alignment, where LBBE captures maximum causal information with minimum background interference.
\end{itemize}

\subsection{Cross-Dataset Transferability}
\label{sec:cross_dataset_default}
We further evaluate whether a single fixed configuration generalizes across datasets without per-dataset tuning. We fix $\lambda_1=\lambda_2=1.0$ and $\tau_1=\tau_2=0.5$, and compare MART, UIAT, DHAT, and HICAT on CIFAR-10, CIFAR-100, SVHN, and MNIST under a unified protocol. The results are summarized in Table~\ref{tab:cross_dataset_unified}.

\begin{table}[t]
\centering
\providecommand{\CrossDatasetTableWidth}{\columnwidth}
\caption{
Cross-dataset robustness under a unified default hyperparameter setting ($\lambda_1=\lambda_2=1.0$, $\tau_1=0.5$, $\tau_2=0.5$).
}
\resizebox{\CrossDatasetTableWidth}{!}{
{
\begin{tabular}{llccc}
\toprule
\textbf{Dataset} & \textbf{Method} & Clean$\uparrow$ & AA$\uparrow$ & Robust Gap$\downarrow$ \\
\midrule
\multirow{4}{*}{\textbf{CIFAR-10}} & MART & 82.99 & 50.67 & 9.52 \\
& UIAT & 82.94 & 52.17 & 7.92 \\
& DHAT & 83.95 & 53.10 & 3.51 \\
& \textbf{HICAT (Ours)} & \textbf{85.92} & \textbf{54.56} & \textbf{2.86} \\
\midrule
\multirow{4}{*}{\textbf{CIFAR-100}} & MART & 54.69 & 27.25 & 9.96 \\
& UIAT & 57.65 & 29.03 & 11.70 \\
& DHAT & 59.14 & 30.17 & 4.24 \\
& \textbf{HICAT (Ours)} & \textbf{62.38} & \textbf{31.24} & \textbf{3.12} \\
\midrule
\multirow{4}{*}{\textbf{SVHN}} & MART & 95.21 & 70.34 & 8.12 \\
& UIAT & 95.16 & 71.84 & 6.65 \\
& DHAT & 96.17 & 72.77 & 4.05 \\
& \textbf{HICAT (Ours)} & \textbf{97.05} & \textbf{74.23} & \textbf{3.18} \\
\midrule
\multirow{4}{*}{\textbf{MNIST}} & MART & 98.34 & 91.27 & 4.68 \\
& UIAT & 98.29 & 92.77 & 3.92 \\
& DHAT & 98.88 & 93.70 & 2.95 \\
& \textbf{HICAT (Ours)} & \textbf{99.22} & \textbf{95.16} & \textbf{2.21} \\
\bottomrule
\end{tabular}
}
}
\label{tab:cross_dataset_unified}
\end{table}

The same default configuration preserves a consistent ranking across the evaluated datasets. Compared with MART, UIAT, and DHAT, HICAT achieves stronger AA robustness while retaining high clean accuracy, suggesting that the gains are not mainly driven by dataset-specific tuning.

\begin{table}[t]
    \begin{center}
     \caption{
        Comparison of per-epoch computational costs (\# RTX4090 GPU hours) using different adversarial training methods on CIFAR-10 under the same software and hardware setup.
    }
    \label{tb:computation}
    \resizebox{\columnwidth}{!}
    {
    {\arrayrulecolor{black}
        \begin{tabular}{l c c c}
            \toprule
            \multicolumn{1}{c}{\multirow{2}{*}{Method}} & \multicolumn{3}{c}{Architecture}\\
            \cmidrule{2-4}
             ~ & ResNet-18 & WRN28-10 & Inception-V3 \\
            \midrule
            TRADES~\citep{TRADES} & 0.024 & 0.161 & 0.188 \\
            MART~\citep{MART}  & 0.026 & 0.168 & 0.194 \\
            AWP~\citep{AWP}    & 0.034 & 0.192 & 0.224 \\
            FSR~\citep{FSR}    & 0.036 & 0.205 & 0.239 \\
            CFA~\citep{CFA}    & 0.031 & 0.184 & 0.213 \\
            UIAT~\citep{UIAT}  & 0.028 & 0.171 & 0.198 \\
            SGLR~\citep{SGLR}  & 0.027 & 0.167 & 0.194 \\
            DHAT~\citep{zhang2025towards} & 0.029 & 0.173 & 0.201 \\
            \textbf{HICAT (Ours)} & 0.029 & 0.176 & 0.204 \\
            \bottomrule
        \end{tabular}}}
    \end{center}
\end{table}

\subsection{Computational Efficiency Analysis}
We further examine the computational overhead introduced by HICAT during adversarial training.
Computational efficiency is important for adversarial training because robust optimization repeatedly generates adversarial examples and is often limited by GPU cost~\citep{shafahi2019adversarial,huang2023fast}.

Table~\ref{tb:computation} reports RTX4090 GPU hours per training epoch for adversarial training on CIFAR-10 under the same schedule, hardware/software setup, and implementation protocol. We include TRADES as a direct low-cost baseline. For alignment-based AT variants (MART, UIAT, DHAT, and HICAT), we use the same runtime setting with parallel inputs; MART is re-run in this unified setting rather than reported only from its original sequential forward implementation. HICAT includes the inverse branch for $\check{x}$ and $\check{z}$, but its branches are evaluated together in one GPU forward pass. Compared with DHAT, the extra operation is only the scalar gate $w(x)$ from the frozen proxy $g_\phi$, which does not require another target-backbone pass.

Table~\ref{tb:computation} accounts for per-epoch adversarial-training cost only. The one-time warm-up cost for training $g_{\phi}(\cdot)$ is reported in Table~\ref{tb:lbbe_warmup_rtx4090}.

\begin{table}[t]
    \centering
    \providecommand{\LBBEWarmupTableWidth}{\columnwidth}
    \caption{One-time warm-up cost for training the proxy model $g_{\phi}(\cdot)$ to approximate the pseudo-label $b(x)$ on a single RTX 4090 GPU (ResNet-18 proxy).}
    \label{tb:lbbe_warmup_rtx4090}
    \resizebox{\LBBEWarmupTableWidth}{!}{
    {
    \begin{tabular}{lccc}
        \toprule
        \textbf{Dataset} & \textbf{Data Used} & \textbf{Proxy Epochs} & \textbf{RTX4090 GPU Hours} \\
        \midrule
        CIFAR-10 & 1k sampled images & 20 & 0.041 \\
        CIFAR-100 & 1k sampled images & 20 & 0.046 \\
        ImageNet-1K & 10k sampled images & 10 & 0.538 \\
        \bottomrule
    \end{tabular}
    }}
\end{table}

\begin{figure*}[t]

    \centering

    \setlength{\tabcolsep}{1pt}

    \renewcommand{\arraystretch}{1}

    \begin{tabular}{@{}ccccc@{}}

        \includegraphics[width=0.19\linewidth]{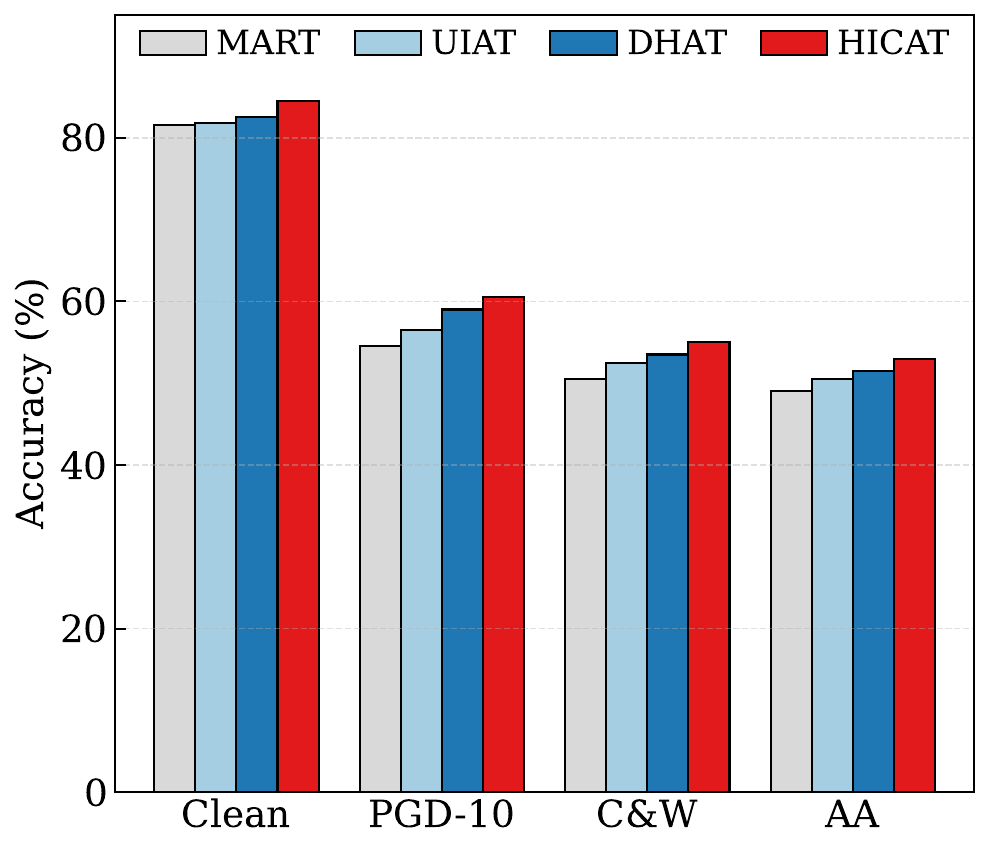} &

        \includegraphics[width=0.19\linewidth]{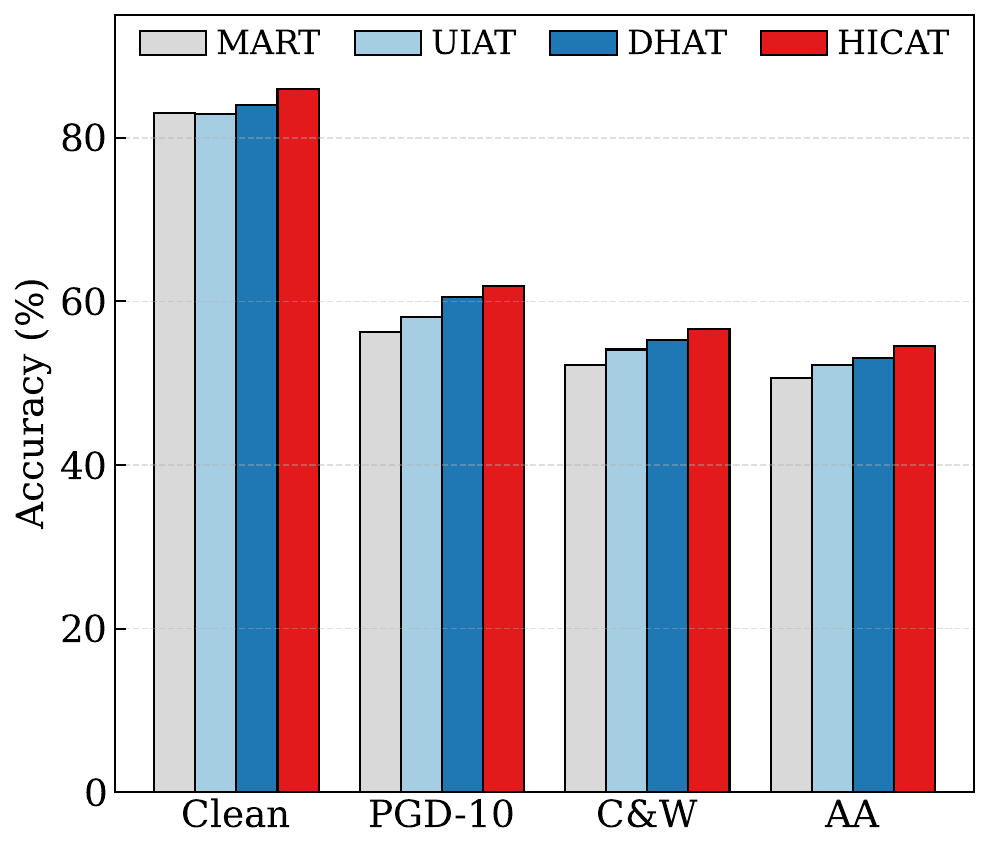} &

        \includegraphics[width=0.19\linewidth]{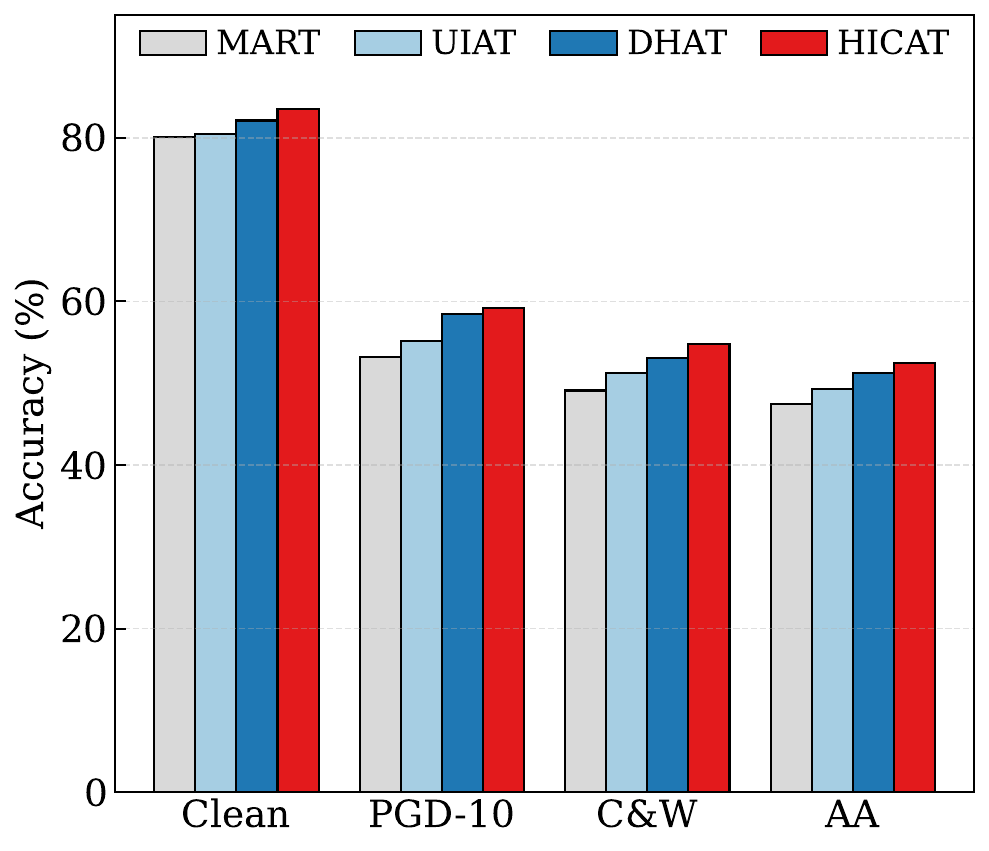} &

        \includegraphics[width=0.19\linewidth]{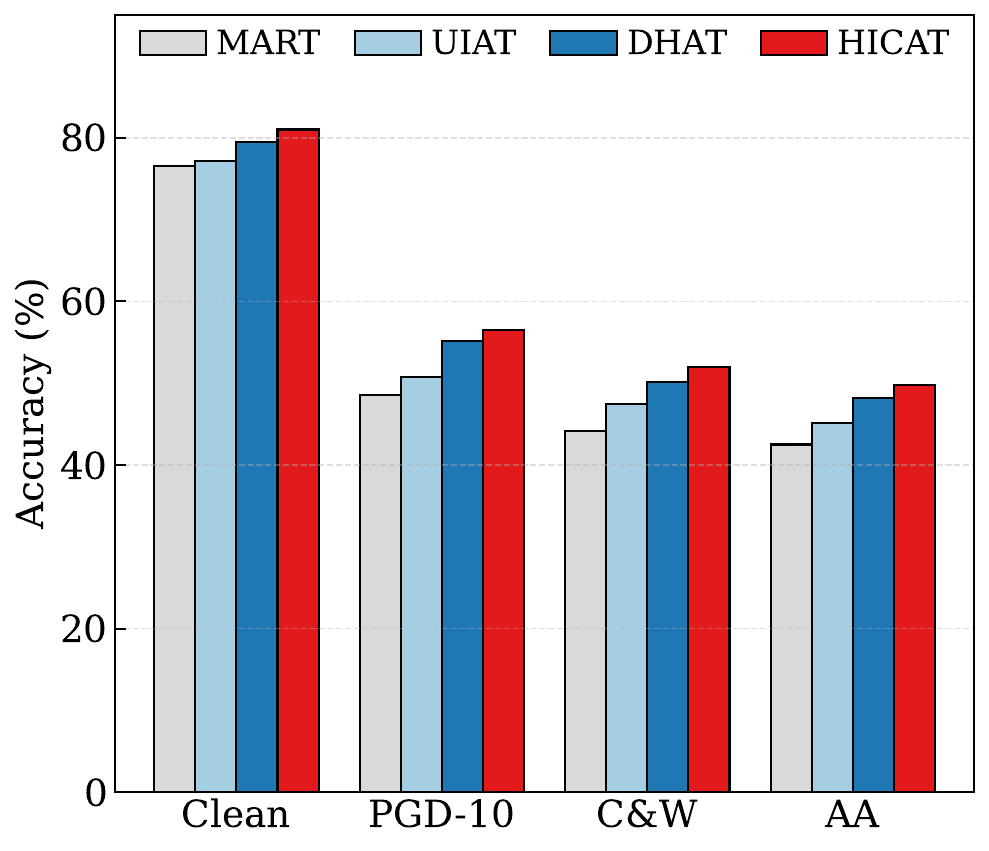} &

        \includegraphics[width=0.19\linewidth]{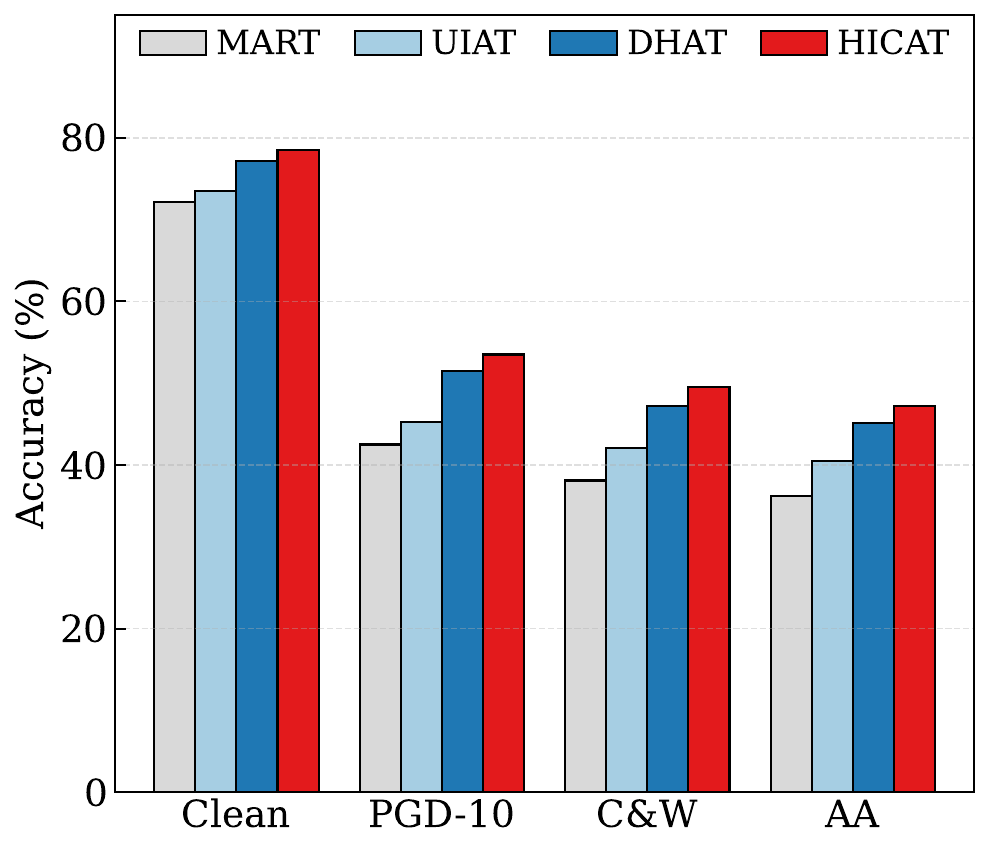} \\

        \footnotesize{(a) $\epsilon_{\text{inv}}=2/255$} &

        \footnotesize{(b) $\epsilon_{\text{inv}}=4/255$} &

        \footnotesize{(c) $\epsilon_{\text{inv}}=8/255$} &

        \footnotesize{(d) $\epsilon_{\text{inv}}=12/255$} &

        \footnotesize{(e) $\epsilon_{\text{inv}}=16/255$}
    \end{tabular}
    \caption{
        \textbf{Impact of inverse adversarial perturbation strength ($\epsilon_{\text{inv}}$) on model performance.}
        We evaluate clean accuracy and robustness (PGD-10, C\&W, AA) on CIFAR-10 across varying training perturbation budgets.
        While baselines (MART, UIAT) suffer from severe performance degradation and overfitting as $\epsilon_{\text{inv}}$ increases (from 8/255 to 16/255),
        \textbf{HICAT (Ours)} demonstrates superior stability, maintaining high robustness even under aggressive inverse parameter settings.
    }
    \label{fig:inverse_eps_ablation}
\end{figure*}

\subsection{Stability to Mask Quality and Generators}
\label{mask:mask_impact}
To investigate the dependence on mask quality, we evaluate HICAT with different mask generators using WRN28-10. The default Grad-CAM masks are produced by a frozen CAM source model, and $g_\phi$ is frozen after warm-up. Therefore, HICAT does not rely on the evolving target model's early-stage attention. Table~\ref{tab:mask_methods} summarizes the results.

\begin{table}[t]
\centering
\providecommand{\MaskMethodTableWidth}{\columnwidth}
\caption{
Robustness (\%) with different mask generators on CIFAR-10 using the WRN28-10 backbone under the same training and evaluation protocol.
}
\resizebox{\MaskMethodTableWidth}{!}{
{\arrayrulecolor{black}
\begin{tabular}{lccc}
\toprule
\textbf{Mask Generator} & Clean$\uparrow$ & AA$\uparrow$ & Robust Gap$\downarrow$ \\
\midrule
Grad-CAM        &  85.92 & 54.56 & 2.86 \\
Grad-CAM++ [1]      & 86.10 & 54.75 & 2.82 \\
Integrated-Grad [2] & 86.38 & 55.08 & 2.57 \\
SOLO [3]       & 86.81 & 55.62 & 2.31 \\
SAM [4]   & 87.42 & 56.10 & 1.95 \\
\bottomrule
\end{tabular}
}
}
\arrayrulecolor{black}
\label{tab:mask_methods}
\end{table}

Table~\ref{tab:mask_methods} shows that replacing Grad-CAM with Grad-CAM++~\cite{chattopadhay2018gradcam++}, Integrated Gradients~\cite{sundararajan2017axiomatic}, SOLO~\cite{wang2020solo}, and SAM~\cite{kirillov2023segment} improves AutoAttack robustness from 54.56\% to 54.75\%, 55.08\%, 55.62\%, and 56.10\%, respectively. The robust gap also decreases from 2.86\% to 1.95\%. These results indicate that better segmentation quality leads to stronger robustness. We use Grad-CAM as the default generator because it has lower computational cost; SAM achieves the highest robustness in this study but requires substantially higher segmentation cost.

\subsection{Impact of Inverse Adversarial Perturbation Strength}
\label{sec:inverse_eps_impact}

We analyze the sensitivity to the inverse perturbation magnitude $\epsilon_{\text{inv}}$, which governs the strength of high-confidence supervision. We conduct a sweep with  $\epsilon_{\text{inv}} \allowbreak \in \{2, 4, 8, 12, 16\}/255$ on CIFAR-10 (Fig.~\ref{fig:inverse_eps_ablation}). Note that MART is evaluated using UIAT's inverse examples to ensure a fair comparison of alignment capabilities.
Three critical trends emerge:

\begin{itemize}
    \item \textbf{Optimal performance at moderate strength.}
    Performance peaks around $\epsilon_{\text{inv}} = 4/255$ for all methods (Fig.~\ref{fig:inverse_eps_ablation}(b)).
    At this sweet spot, inverse examples strike an ideal balance: providing sufficient margin guidance to rectify the decision boundary without destroying input semantics. HICAT achieves its highest AutoAttack accuracy (54.56\%) here, confirming optimal utilization of high-confidence supervision.

    \item \textbf{Baselines suffer from noise amplification.}
    Performance for MART and UIAT degrades sharply beyond $\epsilon_{\text{inv}} \ge 8/255$. Under aggressive perturbation ($\epsilon_{\text{inv}} = 16/255$), UIAT's accuracy collapses to $\sim 40.5\%$.
    This collapse occurs because strong inverse perturbations indiscriminately \textit{amplify non-robust background textures} to boost confidence. Without filtering, baselines blindly align with these intensified artifacts, leading to severe \textit{contextual overfitting} where the model memorizes noise rather than robust features.

    \item \textbf{HICAT achieves stability via dynamic decoupling.}
    In contrast, HICAT maintains robust accuracy over 52\% (AA) even at $\epsilon_{\text{inv}} = 16/255$, outperforming the best baseline by $\sim 7\%$.
    This stability validates {Adaptive Debiasing} as a dynamic filter. As $\epsilon_{\text{inv}}$ increases and background noise becomes more pronounced, LBBE detects the stronger spurious correlation and triggers more aggressive suppression. This effectively \textit{decouples supervision strength from feature corruption}, allowing HICAT to benefit from strong supervision targets without overfitting to exacerbated background noise.
\end{itemize}

\begin{table}[t]
    \centering
    \arrayrulecolor{black}
    \caption{
        \textbf{Impact of LBBE Proxy Architectures on Bias Estimation.}
        We evaluate the sensitivity of HICAT to the proxy backbone used for bias diagnosis.
        \textit{Bias Alignment} measures the similarity between the learned estimator $g_\phi(x)$ and the ground-truth $b(x)$ using \textbf{Pearson Correlation ($r$)} and \textbf{Mean Squared Error (MSE)}.
        UIAT and DHAT are included as baselines. The \textit{Default} HICAT setting uses ResNet-18 as the proxy.
    }
    \resizebox{1.0\linewidth}{!}{
        \begin{tabular}{l|cccc}
            \toprule
            \multirow{2}{*}{\textbf{Method / Proxy}} & \multicolumn{2}{c}{\textbf{Model Performance (\%)}} & \multicolumn{2}{c}{\textbf{Bias Alignment}} \\
            & Clean$\uparrow$ & AA$\uparrow$ & Pearson $r\uparrow$ & MSE$\downarrow$ \\
            \midrule
            UIAT~\citep{UIAT} & 82.94 & 52.17 & - & - \\
            DHAT~\citep{zhang2025towards} & 83.95 & 53.10 & - & - \\
            \midrule
            HICAT (AlexNet) & 84.47 & 54.05 & 0.76 & 0.082 \\
            HICAT (ResNet-18) & 85.92 & 54.56 & 0.89 & 0.035 \\
            HICAT (ResNet-50) & \textbf{86.05} & \textbf{54.68} & \textbf{0.92} & \textbf{0.031} \\
            HICAT (ViT-B/16) & 85.45 & 54.32 & 0.82 & 0.056 \\
            \bottomrule
        \end{tabular}
    }
    \label{tab:proxy_ablation}
\end{table}

\subsection{Robustness of the Bias Estimator}
To verify that our LBBE module captures intrinsic contextual dependencies rather than architectural artifacts, we instantiated the proxy classifier with diverse backbones ranging from shallow CNNs to Transformers. We quantify the estimation quality using the Pearson correlation ($r$) between the predicted and ground-truth bias. As shown in Table~\ref{tab:proxy_ablation}, three key trends emerge:

\begin{itemize}
    \item \textbf{High lower bound over baselines.} Even when utilizing a lightweight \textbf{AlexNet} proxy, HICAT achieves \textbf{54.05\%} AutoAttack accuracy. This significantly outperforms the strongest matched baselines UIAT (52.17\%) and DHAT (53.10\%), confirming that our improvements stem from the \textit{Adaptive Causal Alignment} mechanism itself, rather than reliance on a heavy proxy model.

    \item \textbf{Correlation between diagnosis and robustness.} There is a clear positive correlation between bias alignment and final robustness. As the estimator becomes more precise ($r$ increases from 0.76 with AlexNet to 0.92 with ResNet-50), the robust accuracy consistently improves. This validates our core hypothesis: accurate diagnosis of ``supportive vs. spurious'' context is the decisive factor for robust feature learning.

    \item \textbf{Universality and optimal trade-off.} The framework proves effective across distinct inductive biases, with the Transformer-based \textbf{ViT-B/16} proxy maintaining strong performance ($r=0.82$, 54.32\% AA). Furthermore, while ResNet-50 yields the highest alignment ($r=0.92$), the default \textbf{ResNet-18} strikes the optimal efficiency-accuracy balance, achieving competitive robustness (54.56\% AA) with significantly lower computational overhead.
\end{itemize}

\begin{table}[t]
\centering
\caption{
Ablation study of HICAT components on CIFAR-10 (ResNet-18).
We report clean accuracy (\%), robustness under PGD-10 and AutoAttack (\%),
and the robust generalization gap. When LBBE is removed but CC or FLOE remains active, we set $\hat{b}(x)=0$, yielding $w(x)=0.5$ and FLOE weight $1-|\hat{b}(x)|=1.0$.
Best results are highlighted in \textbf{bold}.
}
\providecommand{\AblationTableWidth}{\linewidth}
\resizebox{\AblationTableWidth}{!}{
{\arrayrulecolor{black}
\begin{tabular}{c c c | c c c c}
\toprule
LBBE & CC & FLOE
& Clean$\uparrow$ & PGD\mbox{-}10$\uparrow$ & AA$\uparrow$ & Gap$\downarrow$ \\
\midrule
 &  &
 & 78.10 & 55.30 & 48.21 & 3.80 \\

\checkmark &  &
 & 78.62 & 55.95 & 48.70 & 2.67 \\

 & \checkmark &
 & 78.85 & 56.42 & 49.18 & 2.43 \\

 &  & \checkmark
 & 78.51 & 56.10 & 49.05 & 2.41 \\

\checkmark & \checkmark &
 & 79.03 & 56.93 & 49.81 & 2.10 \\

\checkmark &  & \checkmark
 & 78.96 & 56.70 & 49.66 & 2.26 \\

\checkmark & \checkmark & \checkmark
& \textbf{79.28} & \textbf{57.21} & \textbf{50.25} & \textbf{2.07} \\
\bottomrule
\end{tabular}}}
\label{tb:hicat_ablation}
\end{table}

\subsection{Ablation Study}

We conduct an ablation on CIFAR-10 (ResNet-18). Table~\ref{tb:hicat_ablation} shows the effect of adding LBBE, CC, and FLOE step by step.
When LBBE is removed but CC or FLOE remains active, we set $\hat{b}(x)=0$. By Eqs.~(\ref{eq:gate_w}) and (\ref{eq:floe_bx}), $w(x)=0.5$ and FLOE weight $1-|\hat{b}(x)|=1.0$. This setting keeps the loss forms unchanged while isolating the benefit of adaptive diagnosis.
Three mechanistic observations emerge:

\begin{itemize}
    \item \textbf{Each alignment branch helps under the fixed bias setting.}
    Activating CC or FLOE with $\hat{b}(x)=0$ already improves over the base model, confirming that both logit rectification and geometric disentanglement contribute to robustness.

    \item \textbf{Adaptive diagnosis further strengthens both branches.}
    Replacing this fixed score with learned LBBE improves AA from 49.18\% to 49.81\% for CC and from 49.05\% to 49.66\% for FLOE, showing that adaptive bias estimation benefits both the logit-space and geometric regularizers.

    \item \textbf{Full HICAT achieves synergistic integration.}
    Combining LBBE, CC, and FLOE yields the highest clean and robust accuracy.
    This validates the ``Measure-Debias-Align'' pipeline: diagnosis (LBBE) requires actionable mechanisms (CC/FLOE), while rigorous constraints require precise diagnosis. The full framework achieves a Causal Alignment where diagnosis and correction reinforce each other.
\end{itemize}

\begin{table}[t]
\centering
\caption{Class-wise weight ablation on CIFAR-10 using the WRN28-10 backbone under $\ell_\infty$ perturbations ($\epsilon=8/255$).}
\label{tb:classwise_lbbe_mean}
\resizebox{\linewidth}{!}{
{\arrayrulecolor{black}
\begin{tabular}{lccccc}
\toprule
\textbf{Method} & Clean$\uparrow$ & PGD-10$\uparrow$ & C\&W$\uparrow$ & AA$\uparrow$ & Robust Gap$\downarrow$ \\
\midrule
HICAT (no adaptive weight) & 83.51 & 60.04 & 54.72 & 52.99 & 4.85 \\
HICAT (class-wise weight) & 84.62 & 61.25 & 55.80 & 53.73 & 3.22 \\
\textbf{HICAT (sample-wise weight, Ours)} & \textbf{85.92} & \textbf{61.87} & \textbf{56.59} & \textbf{54.56} & \textbf{2.86} \\
\bottomrule
\end{tabular}}}
\end{table}

Table~\ref{tb:classwise_lbbe_mean} compares fixed, class-wise, and sample-wise weighting under the same setting. The no-adaptive-weight variant sets $\hat{b}(x)=0$, so CC uses $w(x)=0.5$ and FLOE uses weight $1-|\hat{b}(x)|=1.0$. It obtains 52.99\% AA with a 4.85\% robust gap. Replacing these fixed weights with class-wise weights improves AA to 53.73\% and reduces the gap to 3.22\%, showing that LBBE-based background diagnosis is useful. Sample-wise HICAT further improves AA to 54.56\% and reduces the gap to 2.86\%, showing that per-sample weighting is more precise than assigning one weight to each class.

\section{Conclusion}

In this work, we identified the ``Feature Alignment Paradox'' inherent to adversarial training: visual context is not merely noise but a dual-natured signal that can serve as either a supportive prior or a spurious confounder. Consequently, existing blind suppression strategies in robust learning are limited because they can cause severe semantic degradation.
To resolve this, we proposed \textbf{HICAT}, a unified framework that establishes a Causal Alignment for robust optimization. By integrating the LBBE for diagnosis and FLOE for geometric regularization, HICAT dynamically disentangles causal features from adversarial noise, preserving helpful context while filtering harmful bias.
Empirical results across CIFAR and ImageNet benchmarks demonstrate that HICAT achieves state-of-the-art adversarial robustness. Furthermore, its success across CNNs and ViTs validates \textbf{Causal Alignment} as an architecture-agnostic defense principle that significantly reduces the robust generalization gap.

\bibliographystyle{spbasic}
\bibliography{reference}

\end{document}